%% file: main.tex
\documentclass[journal, preprint]{vgtc}                 

\input{preamble}

\title{GS-ProCams: Gaussian Splatting-Based Projector-Camera Systems}

\author{
  \authororcid{Qingyue Deng}{0009-0001-8471-2480},
  \authororcid{Jijiang Li}{0009-0003-2823-751X},
  \authororcid{Haibin Ling}{0000-0003-4094-8413}, and 
  \authororcid{Bingyao Huang}{0000-0002-8647-5730}
}

\authorfooter{
  \item
    Qingyue Deng is with Southwest University.
    E-mail: qingyue@email.swu.edu.cn.
  \item
    Jijiang Li is with Southwest University.
    E-mail: jijiangli@email.swu.edu.cn.
  \item
  	Haibin Ling is with Westlake University.
  	E-mail: linghaibin@westlake.edu.cn
  \item
    Bingyao Huang is with Southwest University.
    E-mail: bhuang@swu.edu.cn.
    Corresponding author.
}

\abstract{
  We present GS-ProCams, the first Gaussian Splatting-based framework for projector-camera systems (ProCams). GS-ProCams is not only view-agnostic but also significantly enhances the efficiency of projection mapping (PM) that requires establishing geometric and radiometric mappings between the projector and the camera. Previous CNN-based ProCams are constrained to a specific viewpoint, limiting their applicability to novel perspectives. In contrast, NeRF-based ProCams support view-agnostic projection mapping, however, they require an additional co-located light source and demand significant computational and memory resources. To address this issue, we propose GS-ProCams that employs 2D Gaussian for scene representations, and enables efficient view-agnostic ProCams applications. In particular, we explicitly model the complex geometric and photometric mappings of ProCams using projector responses, the projection surface's geometry and materials represented by Gaussians, and the global illumination component. Then, we employ differentiable physically-based rendering to jointly estimate them from captured multi-view projections. Compared to state-of-the-art NeRF-based methods, our GS-ProCams eliminates the need for additional devices, achieving superior ProCams simulation quality. It also uses only 1/10 of the GPU memory for training and is 900 times faster in inference speed. Please refer to our project page for the code and dataset: \url{https://realqingyue.github.io/GS-ProCams/}.
}

\keywords{Projection mapping, projector-camera systems, gaussian splatting.}

\teaser{
  \centering
  \includegraphics[width=\linewidth]{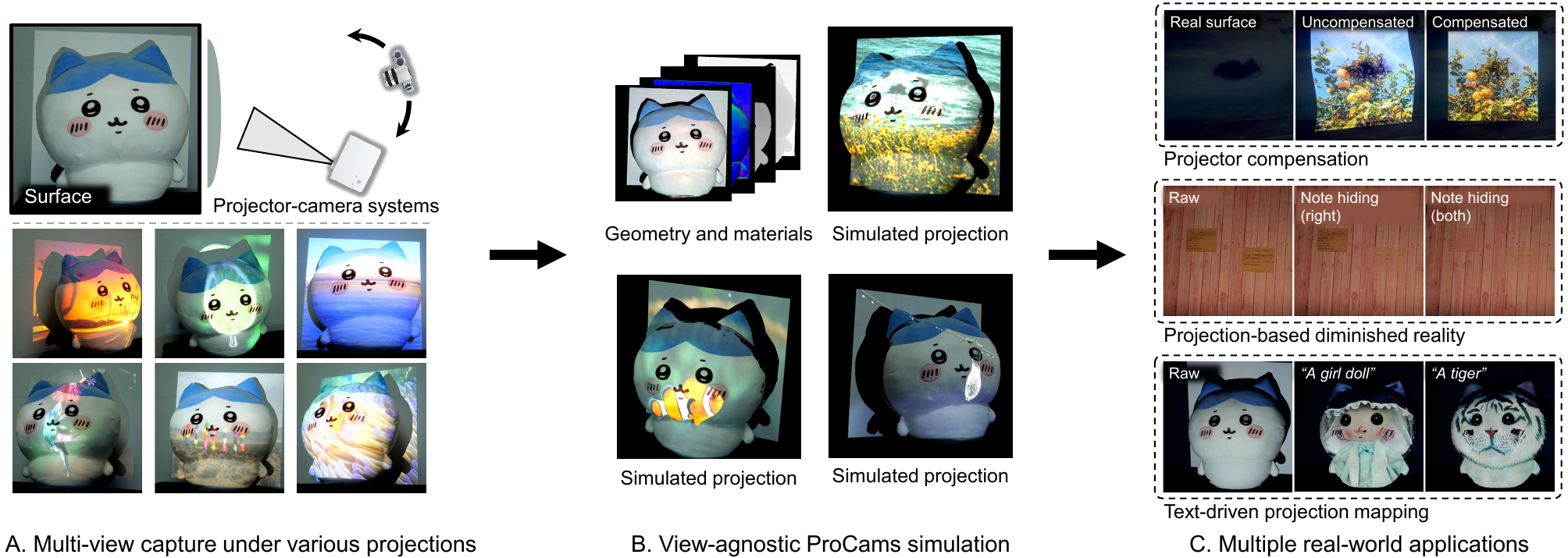}
    \caption{
      \textbf{GS-ProCams.} We represent the scene using 2D Gaussian, allowing us to efficiently model the geometric and photometric mappings of ProCams. \textbf{(A)} By projecting various patterns onto the projection surface and capturing images from multiple viewpoints using a camera, \textbf{(B)} we achieve high-fidelity ProCams simulation while estimating the projection surface's geometry and materials. \textbf{(C)} After training on a scene, GS-ProCams can achieve efficient view-agnostic projector compensation, projection-based diminished reality~\cite{Iwai2006DR, mori2017DRsurvey} (with inpainting models~\cite{suvorov2022lama}), and projection mapping~\cite{Iwai2024PMReview} (with diffusion models~\cite{rombach2022diffusion, hachnochi2023cdc}).} 
  \label{fig: teaser}
}

\begin{document}

\firstsection{Introduction}
\maketitle
\label{sec:intro}
Projector-camera systems (ProCams) are integrated setups that combine one or more projectors and cameras, and have emerged as indispensable tools in computer vision~\cite{shandilya2023NeRFSL, Mirdehghan2024TurboSL, Morgenstern2023X-Maps, Shin2024DispersedSL}, and in spatial augmented reality (SAR)/projection mapping (PM)~\cite{Iwai2024PMReview, Takeuchi-Iwai-2024PMwithEnvLight, Toshiyuki2025BRDFDiffuison}. ProCams have enabled exceptional experiences in immersive interaction~\cite{Duan2018FlyingHand, Hartman2020AAR,Yuri-Watanbabe-2022DPM,Sato-Iwai-2024Hand, erel2024casper}, artistic creation and exhibition~\cite{SCHMIDT2019Museum, Peng-Watanbabe-2023Face, Yasui-Watanbabe-2024ProjectionMW, Deng2025LAPIG}, as well as industrial design and manufacturing~\cite{Takezawa-Iwai-2019MaterialSR}, among other domains. Alongside these developments, recent advances in image generation~\cite{rombach2022diffusion, suvorov2022lama, Zhang2024DiffMorpher} have introduced powerful capabilities for projection mapping. These techniques enable ProCams to interactively and convincingly alter the perceived appearance of real-world objects, thereby significantly expanding the expressive and functional scope of ProCams-based applications.

At the core of ProCams-based applications lie geometric and photometric mappings between the projectors and cameras. Structured light utilizes the geometric and photometric constraints of ProCams to estimate the object geometry and materials~\cite{sansoni2009state, salvi2010state, Geng20113DImageAcqui, Chunyu2019ProCamssfM}. Furthermore, through geometric registration~\cite{Boroomand2016Saliency, Ueno2022SolidModeling, Raskar2000Immersive, Yamazaki2011SimuCalib, Willi2017RobustGeoCalib} and photometric compensation~\cite{WetzsteinBimber2007radiometric, Kurth2020AdapColorCorrec, Grundhofer-Iwai-2015PhotometricPC, Ashdown2006RobustPPC, Huang2019Compenet, nayar2003projection}, ProCams can achieve full projector compensation to rectify projections on non-planar and textured surfaces, ensuring high-fidelity visual results.

Despite significant efforts to model projector-surface interactions under the assumption of a static spatial relationship between the projector and the projection surface, most existing methods~\cite{Kageyama-Iwai-2024DistortionFree, Huang2021DeProcams, Kageyama-Iwai-2022OnlineDeblur, Huang2022CompenNeSt++, Wang2023CompenHR, Wang2024Vicomp, Li2025DPCS} establish geometric and photometric correspondences that are tightly coupled to a specific viewpoint. As a result, these learned or computed relationships fail to generalize when the viewpoint changes. The neural field-based framework~\cite{Erel2023Nepmap} utilizes multilayer perceptrons (MLPs)~\cite{Boss2021Neural-PIL, Srinivasan2021NeRV, Zhang2021NeRFactor} to construct an integrated neural representation for view-agnostic ProCams simulation. While it disentangles objects' geometry and materials through a multi-stage training strategy and supports view-agnostic projector compensation, this method requires an additional co-located light and a dark room (i.e., assumes no environment light). Moreover, it suffers from low computational and memory efficiency, which limits its applicability in real-world scenarios.

We propose GS-ProCams, an efficient and fully differentiable framework for view-agnostic projection mapping, enabling practical applications with non-static observers, as illustrated in~\cref{fig: teaser}. A key challenge in this domain is modeling the complex interaction between the projector and real-world surfaces. To address this, we utilize 2D Gaussians~\cite{Huang20242DGS} as the ProCams scene representation. Since the original 2D Gaussians do not explicitly capture the Bidirectional Reflectance Distribution Function (BRDF), we augment each Gaussian point with BRDF attributes. To support dynamic, high-resolution projection patterns, we first derive surface attributes from Gaussians and then feed them into a unified lighting computation, where a compact convolution kernel is jointly optimized to capture the projector's Point Spread Function (PSF). We further leverage the anisotropic expressiveness of 2D Gaussians to approximate residual global illumination as a view-dependent residual color, effectively capturing environment illumination and enhancing applicability. Rather than merely combining existing techniques, our framework is designed with a unified representation and optimization objective that couples the projector responses, scene geometry, material attributes, and residual illumination into a single differentiable system. 
Experiments on both ProCams simulation and projector compensation show clear advantages of our GS-ProCams over previous methods. 

In summary, our primary contributions include:
\begin{itemize}
\item We present GS-ProCams, the first Gaussian Splatting-based framework for projector-camera systems. It integrates 2D Gaussians and physically-based rendering to efficiently model the geometric and photometric mappings of ProCams.
\item Compared to NeRF-based ProCams, GS-ProCams eliminates the need for additional devices, can operate under ambient room lighting, and is 900 times faster in inference speed while using only 1/10 of the GPU memory for training.
\item GS-ProCams can be applied to multiple view-agnostic ProCams applications simultaneously, including projection mapping and projector compensation.
\item We introduce a view-agnostic ProCams benchmark encompassing various viewpoints, textured surfaces, and environmental lighting conditions for training and evaluating ProCams models. This benchmark is designed to foster future research in the field.
\end{itemize}

\section{Related Work}
\subsection{ProCams simulation}
ProCams simulation aims to model the real-world photometric and geometric mappings of projector-camera systems (ProCams) within a computational space. Once these mappings are established, the simulation synthesizes camera-captured images corresponding to various projections. ProCams simulation facilitates image-based relighting~\cite{WetzsteinBimber2007radiometric, Huang2021DeProcams, Erel2023Nepmap, Li2025DPCS}, shape reconstruction~\cite{shandilya2023NeRFSL, Mirdehghan2024TurboSL, Huang2021DeProcams, Erel2023Nepmap}, and projector compensation~\cite{Kageyama-Iwai-2022OnlineDeblur, Wang2023CompenHR}. Early methods describe the pixel mapping from projectors to cameras using the Light Transport Matrix (LTM)~\cite{Sen2005DualPhotography, O2010OpticalLTM, O'Toole2014StructuredLT, WetzsteinBimber2007radiometric}. However, the reconstruction quality of LTM, which is based on the linear relationship between pixels, generally depends on dense sampling and specialized computational designs of large matrices~\cite{debevec2000acquiring, Debevec2000AcquiringTR, Sen2005DualPhotography, WetzsteinBimber2007radiometric, O2010OpticalLTM, Wang2009KernelLTM}. 

DeProCams~\cite{Huang2021DeProcams} generates photorealistic simulations of ProCams using fewer image samples by incorporating inherent epipolar geometric and simplified photometric constraints under a fixed viewpoint with neural rendering based on a convolutional neural network~(CNN). However, a notable limitation of this approach stems from its view-specific design, which precludes the use of multi-view geometric constraints and necessitates manual ProCams calibration with a checkerboard pattern~\cite{Huang2021Calibrate}. Although Park et al.~\cite{Park2022PCFUDR} examine the geometric and photometric mappings in ProCams using a differentiable rendering scheme, it remains limited to specific view information and relies on an RGB-D camera. Additionally, the method depends heavily on pixel-level bias corrections. Li et al.~\cite{Li2025DPCS} model physical ProCams using interpretable parameters based on path tracing to enable high-fidelity lighting effects, and benefit downstream ProCams tasks. However, it relies on geometry reconstructed from a specific viewpoint by gray-coded structured light (SL)~\cite{Moreno2012Calib}, and is thus sensitive to the quality of this reconstruction, making it unsuitable for direct application to novel viewpoints. 

Modeling view-agnostic ProCams remains a challenging task. It requires capturing the geometry of the projection surface and modeling view-dependent reflectance effects, as most real-world surfaces exhibit non-Lambertian properties. Moreover, environmental illumination is often non-negligible in practical indoor environments. For view-specific ProCams, the influence of environment light on the projection surface can be easily accounted for by capturing an image from the scene under environment light only at the same viewpoint. This can be achieved by projecting a plain black image, and the final appearance of the projection surface can then be approximated by linearly combining this with the radiance induced by the projected content. However, for novel viewpoints where synthesis is required, such information is not available beforehand. Erel et al.~\cite{Erel2023Nepmap} integrate the projector into a neural reflectance field~\cite{Boss2021Neural-PIL, Srinivasan2021NeRV, Zhang2021NeRFactor} to enable view-agnostic projection mapping within 3D scenes. Despite this breakthrough~\cite{Erel2023Nepmap}, the influence of other light sources in the environment is ignored, and the approach requires a light source co-located with the camera, which limits its applicability. In~\cref{tab:ProCams_works}, we detail the features and requirements of current ProCams simulation approaches to allow for a clear comparison with our GS-ProCams.

\begin{table}[tb]
    \caption{\textbf{Comparison of representative approaches for ProCams and our GS-ProCams.} We use \fillcolor{red}{\ding{55}} and \fillcolor{green}{\ding{51}} to indicate the limitations and capabilities of different methods, respectively. Notably, view-specific methods can account for environmental illumination by capturing the scene under environment light only at the corresponding viewpoint. However, such information is not available in advance for unseen novel viewpoints required by view-agnostic ProCams. We use \fillcolor{yellow}{\ding{51}} highlights to emphasize this distinction. GS-ProCams is a view-agnostic method that enables multi-view applications. It simultaneously handles both geometric and photometric mappings under environment light, using only an RGB camera and a projector. }
     \label{tab:ProCams_works}
    \centering
    \resizebox{\linewidth}{!}{  
        \renewcommand{\arraystretch}{1.25}
        \begin{tabular}{|l|c|c|c|c|}
            \hline
            \textbf{Method} & View-agnostic & W/o manual & W/o extra & Environment \\
            & method & calibration & devices & light  \\
            \hline
            LTM~\cite{WetzsteinBimber2007radiometric} & \bgred{\ding{55}} & \bggreen{\ding{51}} & \bggreen{\ding{51}} & \bgyellow{\ding{51}}  \\
            CNN-based~\cite{Huang2021DeProcams} & \bgred{\ding{55}} & \bgred{\ding{55}} & \bggreen{\ding{51}} & \bgyellow{\ding{51}}  \\
            Depth-prior~\cite{Park2022PCFUDR} & \bgred{\ding{55}} & \bggreen{\ding{51}} & \bgred{\ding{55}} & \bgyellow{\ding{51}}  \\
            Path-tracing~\cite{Li2025DPCS} & \bgred{\ding{55}} & \bgred{\ding{55}} & \bggreen{\ding{51}} & \bgyellow{\ding{51}} \\
            NeRF-based~\cite{Erel2023Nepmap} & \bggreen{\ding{51}}  & \bggreen{\ding{51}} & \bgred{\ding{55}} & \bgred{\ding{55}}  \\
            GS-ProCams (Ours) & \bggreen{\ding{51}} & \bggreen{\ding{51}} & \bggreen{\ding{51}} & \bggreen{\ding{51}} \\
            \hline
        \end{tabular}
    } 
\end{table}

\subsection{Projector compensation}
Projector compensation adapts projected images to non-planar and textured surfaces to achieve the desired visual appearance. It typically involves geometric~\cite{Boroomand2016Saliency, Ueno2022SolidModeling, Raskar2000Immersive, Yamazaki2011SimuCalib, Willi2017RobustGeoCalib} and photometric~\cite{Grundhofer-Iwai-2015PhotometricPC, Kurth2020AdapColorCorrec, Ashdown2006RobustPPC, Huang2019Compenet, nayar2003projection} compensation, which have been studied separately in prior work.
End-to-end methods enable projector compensation by integrating geometric correction estimation represented by the warping grid~\cite{Huang2022CompenNeSt++, Wang2023CompenHR} or optical flow~\cite{Wang2024Vicomp, Kageyama-Iwai-2024DistortionFree, Li2024PBECompen} with CNN-based photometric compensation modules. While recent work~\cite{Wang2023CompenHR} improves computational and memory efficiency for high-resolution projections, CNNs still struggle to preserve high-frequency details due to spectral bias~\cite{Rahaman2019SpectralBiasNN}, often resulting in oversmoothed outputs. Furthermore, these methods are typically constrained to a specific viewpoint and smooth, uniform surfaces.

An alternative strategy is to leverage ProCams simulation as a forward model to guide projector compensation. Traditional approaches~\cite{WetzsteinBimber2007radiometric} obtain the compensated projector input pattern by applying the inverse LTM. More recently, fully differentiable ProCams~\cite{Huang2021DeProcams, Park2022PCFUDR, Erel2023Nepmap, Li2025DPCS} generate projection patterns by making the synthesized scene closely approximate the desired appearances. Erel et al.~\cite{Erel2023Nepmap} achieve projector compensation from novel viewpoints by view-agnostic ProCams simulation using a neural reflectance field~\cite{Boss2021Neural-PIL, Srinivasan2021NeRV, Zhang2021NeRFactor}. Despite that, this method, which relies on ray-casting and MLPs, poses challenges to computational and memory resources. Additionally, this method assumes no environment light in the scene.

\subsection{Gaussian splatting and inverse rendering}
3D Gaussian Splatting (3DGS)~\cite{Kerbl20233DGS} utilizes 3D Gaussians to represent scenes, enabling multi-view photorealistic image rendering through rasterization techniques. It is renowned for its real-time differentiable rendering capabilities while maintaining rendering quality that rivals leading methods in novel view synthesis~\cite{Mildenhall2021NeRF, Barron2022Mip-NeRF, Xu2022Point-NeRF, Nvidia2022Instant-NGP, Fridovich-Keil2022Plenoxels}. Additionally, recent advancements extend 3DGS to estimate materials and environment lighting within an inverse rendering framework~\cite{Jiang2024GaussianShader, Gao2023Relightable3DGS, Liang2024GS-IR}.

Although 3DGS~\cite{Kerbl20233DGS} is optimized for efficient image rendering, extracting precise geometrical details from Gaussian points is challenging~\cite{Huang20242DGS, chen2023neusg, yu2024gsdf, Guedon2024SuGaR, yu2024GOF, turkulainen2024DN-Splatter}. Conversely, 2D Gaussian Splatting (2DGS)~\cite{Huang20242DGS} employs 2D Gaussian primitives to ensure geometric consistency across multiple viewpoints, but does not explicitly model materials and lighting. We extend 2D Gaussian points by incorporating BRDF parameterization to simulate the interaction between projector light and surfaces in ProCams. 

\begin{figure}[tb]
 \centering 
 \includegraphics[width=\columnwidth]{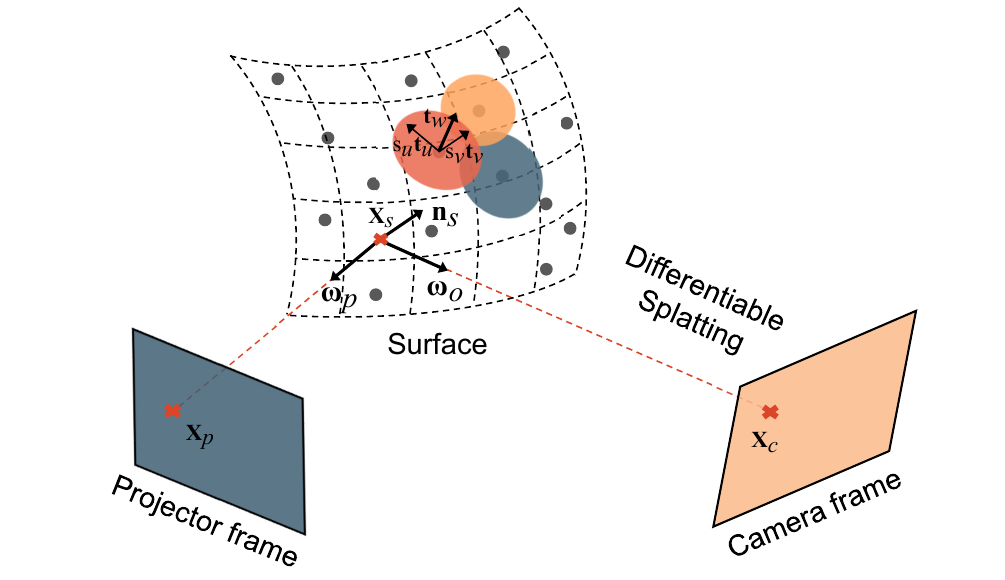}
 \caption{\textbf{GS-ProCams setup.} The intersection point $\mathbf{x}_s$ of the ray from the camera and the projection surface is determined by 2D Gaussians. The direct light from the projector illuminates this point.}
 \label{fig:gsprocams}
\end{figure}

\begin{figure*}[ht]
 \centering 
 \includegraphics[width=\linewidth]{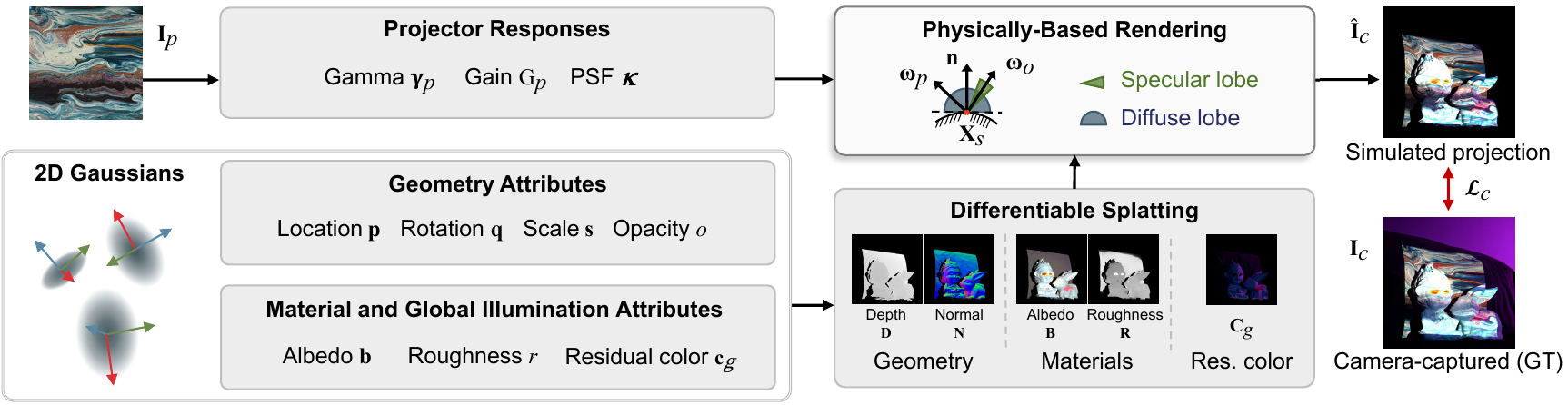}
 \caption{\textbf{GS-ProCams training pipeline.} GS-ProCams uses 2D Gaussians to represent the scene, then obtains the geometry and materials of the projection surface, as well as the global illumination component, through differentiable splatting techniques~\cite{Huang20242DGS}. By using various projection patterns and capturing images from different viewpoints, GS-ProCams jointly optimizes the explicit projector responses and attributes of the 2D Gaussian points listed on the left through physically-based differentiable rendering.}
 \label{fig: pipeline}
\end{figure*}

\section{GS-ProCams} 
\subsection{Preliminary}
\label{sec: preliminary_2dgs}
2DGS~\cite{Huang20242DGS} employs explicit 2D Gaussians to represent scene geometry. The following primary components distinguish each 2D Gaussian point: its central point $\mathbf{p}_k$, two principal tangential vectors $\mathbf{t}_u$ and $\mathbf{t}_v$, and a scaling vector $\mathbf{s} = (s_u, s_v)$ which regulates the variances of the 2D Gaussian distribution. Consequently, 2D Gaussians inherently include inherent normals defined by the above tangential vectors as $\mathbf{t}_w = \mathbf{t}_u \times \mathbf{t}_v $. In world space, every 2D Gaussian is defined in a local tangent plane and parameterized as:
\begin{equation}
\label{eqn:2DGS}
{P(u, v) = \mathbf{p}_k + s_u \mathbf{t}_u u + s_v \mathbf{t}_v v.}
\end{equation}
Then, the 2D Gaussian value for a point $\mathbf{u} = (u, v)$ in its local space can be evaluated by the standard Gaussian:
\begin{equation}
\label{eqn:GS}
G(\mathbf{u}) = \text{exp}\Big(-\frac{u^2+v^2}{2}\Big).
\end{equation}
Additionally, each 2D Gaussian primitive possesses an opacity $o$ and a view-dependent appearance $\mathbf{c}$ represented using spherical harmonics (SH). Finally, given the viewpoint-based color $\mathbf{c}_i$ of each Gaussian point, the pixel color $\mathbf{C}$ is obtained through alpha blending with $K$ tile-based Gaussians arranged according to the depth values of their center locations:
\begin{equation}
\label{eqn:alpha_blending}
\mathbf{C}(\mathbf{x}) = \sum_{i \in K} W_i c_i, \;\; W_i = \alpha_i\prod_{j=1}^{i-1} (1 - \alpha_j),
\end{equation}
where $\alpha = o \cdot \hat{G}(u(\mathbf{x}))$. Here, $\hat{G}(u(\mathbf{x}))$ is the result of applying a low-pass filter to ${G}(u(\mathbf{x}))$, and $\mathbf{x}$ denotes a homogeneous ray that originates from the camera and traverses the UV space.

The orientation of a 2D Gaussian can be constructed as a rotation matrix $\mathbf{R} = [\mathbf{t}_u, \mathbf{t}_v, \mathbf{t}_w]$ and represented as a unit quaternion $\mathbf{q}$ during optimization. In summary, each 2D Gaussian is parameterized by several learnable parameters as $\{\mathbf{p}, \mathbf{q}, \mathbf{s}, o, \mathbf{c}\}$. Furthermore, the designed density control mechanism enables adaptive regulation of the total number of points and the density within a unit volume.

\subsection{ProCams modeling}
\label{sec: ProCams modeling}
The core of our GS-ProCams is establishing the geometric and photometric mappings of projector-camera systems (ProCams) using Gaussian Splatting (GS), thus achieving view-agnostic and high-fidelity ProCams simulation. As illustrated in \cref{fig:gsprocams}, consider a point $\mathbf{x}_s$ on the projection surface within a scene, which is captured by the camera as $\mathbf{x}_c$ in the camera frame. In the absence of occlusion, this point is illuminated by the pixel $\mathbf{x}_p$ in the projector frame. Furthermore, $\mathbf{x}_s$ is also illuminated by environment and complex indirect lighting, i.e., global illumination. Taking these factors into consideration, we utilize the classic rendering equation~\cite{Kajiya1986TheRenderingEqu} to describe the relationship between outgoing and incoming light at point $\mathbf{x}_s$ as follows:
\begin{equation}
\label{eqn:TheRenderEq}
L_o(\omega_o, \mathbf{x}_s) = \int_{\Omega} f(\omega_o, \omega_i, \mathbf{x}_s)L_i(\omega_i, \mathbf{x}_s)(\omega_i \cdot \mathbf{n}_s)d\omega_i,
\end{equation}
where $\Omega$ represents the upper hemisphere centered at $\mathbf{x_s}$, with the normal vector $\mathbf{n}_s$, and $f$ denotes the BRDF. We employ 2D Gaussians described in~\cref{sec: preliminary_2dgs} as foundational primitives, extending their original attributes by BRDF parameterization to capture the interactions between the projector's direct light and the projection surface. As the projector's direct illumination dominates the scene, we estimate a view-dependent residual color term to account for global illumination, as detailed in \cref{sec: global_illumination}, avoiding the computational cost of full global illumination modeling as in DPCS~\cite{Li2025DPCS}. Building upon this, we capture a series of images of the surface $\{\mathbf{I}_c^m\}_{m=1}^M$ from multiple viewpoints, which is illuminated by known projector input patterns $\{\mathbf{I}_p^n\}_{n=1}^N$. Then, we simultaneously estimate the surface's unknown geometric and material properties utilizing physically-based rendering (PBR). We illustrate the pipeline of our GS-ProCams in \cref{fig: pipeline}.

\subsection{Attributes from Gaussians} 
As shown in \cref{fig: pipeline}, we first derive attributes of the projection surface from 2D Gaussians positioned in 3D space. The attributes can generally be divided into three categories: (1) the depth and inherent normals from geometry attributes of Gaussians; (2) the BRDF attributes for material reflectance; and (3) the global illumination component, approximated by view-dependent RGB residual color. Notably, we utilize normals derived from depth gradients in the shading computations, while using inherent normals to regularize geometric consistency.

\label{sec: Attributes}
\noindent\textbf{Depth.} The intersections between camera rays and the surface are determined by their depths in the camera frame. We use the same approach as 2DGS~\cite{Huang20242DGS} and render the mean depth $D$ by \cref{eqn:alpha_blending} to represent the ray-surface intersection as follows:
\begin{equation}
\label{eqn:5:Mean Depth}
D = \sum_{i \in K} \frac{W_i z_i}{\sum_{i \in K}W_i},
\end{equation}
where $z_i$ denotes the depth of the intersection between the homogeneous ray $\mathbf{x}$ corresponding to the camera pixel $\mathbf{x}_c = (x, y)$ and the $i$-th Gaussian planes. 

\noindent\textbf{Normal.} To enhance depth consistency and surface smoothness, we utilize normals derived from viewpoint depth rather than the inherent normals aligned with the directions of their highest density gradients as the normal map $\mathbf{N}$:
\begin{equation}
\label{eqn:Depth_Normal}
\mathbf{N}(x, y) = \frac{\nabla_x \mathbf{d} \times \nabla_y \mathbf{d}}{\left| \nabla_x \mathbf{d} \times \nabla_y \mathbf{d} \right|},
\end{equation}
where x and y denote the coordinates of the pixel $\mathbf{x}_c$ in the depth map $D$. 

\noindent\textbf{BRDF attributes.} To enable the light-surface interactions for the projector, we augment each original 2D Gaussian point~\cite{Huang20242DGS} with additional learnable attributes: albedo $\mathbf{b} \in [0, 1]^3$ and roughness $r \in [0, 1]$. Then we splat these attributes onto the camera frame by utilizing the efficient differentiable rasterization method described by \cref{eqn:alpha_blending}:
\begin{equation}
\label{eqn:BRDF_alpha_blending}
\{\mathbf{B}, R\} = \sum_{i \in K} W_i\{\mathbf{b}_i, r_i\},
\end{equation}
where $\mathbf{B}$ and $R$ represent the maps of albedo and roughness after rasterization in a specified viewpoint.

\subsection{Lighting}
In ProCams settings, light usually comes from the environment and the projector. The projector pixels act like high-power spotlight light sources that dominate the light contribution. Therefore, we model the incident light at $\mathbf{x}_s$ as the sum of projector direct light, denoted as $L_{p}$, and the global illumination, denoted as $L_{g}$.

\vspace{.5em}
\noindent\textbf{Projector direct light.} 
\label{sec: projector direct light}
Denote the projector gamma as $\gamma_{p}$, which converts standard sRGB color into linear RGB space. Denote the projector's luminous power as $G_p$, and assume constant light attenuation. Additionally, due to the projector's optical characteristics, e.g., point spread function (PSF), the projected light suffers from defocus. Therefore, we use a learnable $5 \times 5$ kernel $\kappa$ to simply approximate the projector's PSF. Consequently, the projector emitted direct light $L_{p}$ can be expressed as:
\begin{equation}
\label{eqn:Lprj}
  L_{p}(\mathbf{x}_s) = \kappa \ast \big(G_p \cdot \mathbf{I}_p^{\gamma_p}(\mathbf{x}_p)\big), 
\end{equation} 
where $\mathbf{x}_p$ is the projector pixel that directly illuminates the surface point $\mathbf{x}_s$, as shown in~\cref{fig:gsprocams}, and $\ast$ is the convolution operator. Theoretically, the PSF varies spatially in the image plane~\cite{Kageyama-Iwai-2022OnlineDeblur, Kusuyama-Iwai-2024Shadow-and-deblur}, but accurately measuring or modeling this variation is challenging and computationally intensive. For efficiency, we use a spatially-invariant approximation, with some sacrifice in accuracy.

Denote the projector projection matrix as $\mathbf{M}_p$, the relationship between $\mathbf{x}_p$ and $\mathbf{x}_s$ is given by 
\begin{equation}
\label{eqn:x_s2x_p}
\overline{\mathbf{x}}_p = \mathbf{M}_p \overline{\mathbf{x}}_s,
\end{equation}
where $\overline{\mathbf{x}}_p$ and $\overline{\mathbf{x}}_s$ denote the homogeneous coordinates of $\mathbf{x}_p$ and $\mathbf{x}_s$, respectively. Then, according to \cref{eqn:TheRenderEq}, the outgoing/reflected projector direct light at $\mathbf{x}_s$ can be computed by:
\begin{equation}
\label{eqn:Cprj}
\mathbf{C}_{p}(\omega_o, \mathbf{x}_s) = f(\omega_o, \omega_p, \mathbf{x}_s)L_{p}( \omega_p, \mathbf{x}_s)(\omega_p \cdot \mathbf{n}_s),
\end{equation}
where the normal $\mathbf{n}_s$ of the surface point $\mathbf{x}_s$ is determined by \cref{eqn:Depth_Normal}. We approximate the BRDF $f$ in \cref{eqn:Cprj} using a simplified Disney BRDF model~\cite{burley2012PBRDisney}.

\vspace{.5em}
\noindent\textbf{Global illumination.}
\label{sec: global_illumination}
In most projection mapping settings, direct projector illumination dominates surface appearance and is explicitly modeled in our framework. Meanwhile, global illumination includes environment lighting and weaker projector-induced indirect effects, which also contribute to the final appearance. Although projector-induced indirect effects are input-dependent, accurately modeling them is computationally expensive~\cite{Li2025DPCS}. Inspired by GS-based radiance fields~\cite{Kerbl20233DGS, Huang20242DGS}, we approximate residual global illumination as a view-dependent residual color $\mathbf{c}_g(\omega_o)$ by maintaining spherical harmonics coefficients for each Gaussian point, which are then rasterized by:
\begin{equation}
\label{eqn: Cres_alpha_blending}
\mathbf{C}_{g} = \sum_{i \in K} W_i \mathbf{c}_{g_i}.
\end{equation}

To mitigate degeneracy during joint optimization, we incorporate a plain black projection at training viewpoints to constrain the residual term. This residual primarily captures environmental illumination, while projector-induced indirect effects are coarsely approximated and implicitly absorbed into this term and other components such as albedo during joint optimization. This formulation also supports rendering beyond the projection surface, although performance remains limited in textureless regions of the surrounding scene, which is a known challenge for GS-based radiance methods. Finally, the camera-captured color is given by 
\begin{equation}
\label{eqn:Camera}
\hat{\mathbf{I}}_c = \text{clamp}\left( (\mathbf{C}_{p} + \mathbf{C}_{g})^{{\gamma_c}}, 0, 1\right),
\end{equation}
where $\gamma_c$ is a gamma tone mapping function~\cite{anderson1996sRGB}. 

\begin{figure*}[tb]
 \centering 
 \includegraphics[width=\linewidth]{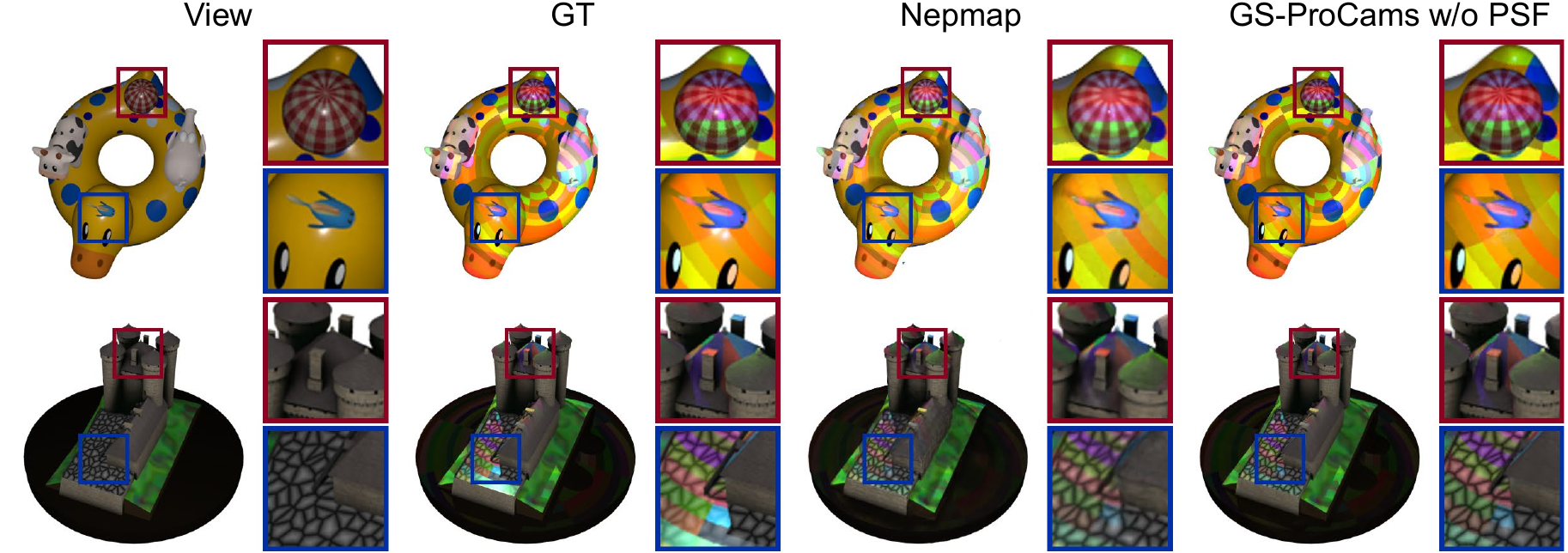}
     \caption{\textbf{Visual comparisons of ProCams simulation on the Nepmap synthetic dataset~\cite{Erel2023Nepmap}.} The first column displays an object from a novel viewpoint, the second column shows the object under a novel projection pattern, and the third and fourth columns present the results of two methods. We present two of the four scenes here as examples. Compared to Nepmap, our model exhibits finer details and more realistic colors. Moreover, our method outperforms Nepmap in computation and memory efficiency by a significant margin (\autoref{tab: synthetic}). See supplementary material for more results.}
 \label{fig: synthetic}
\end{figure*}

\begin{figure}[tb]
 \centering 
 \includegraphics[width=\columnwidth]{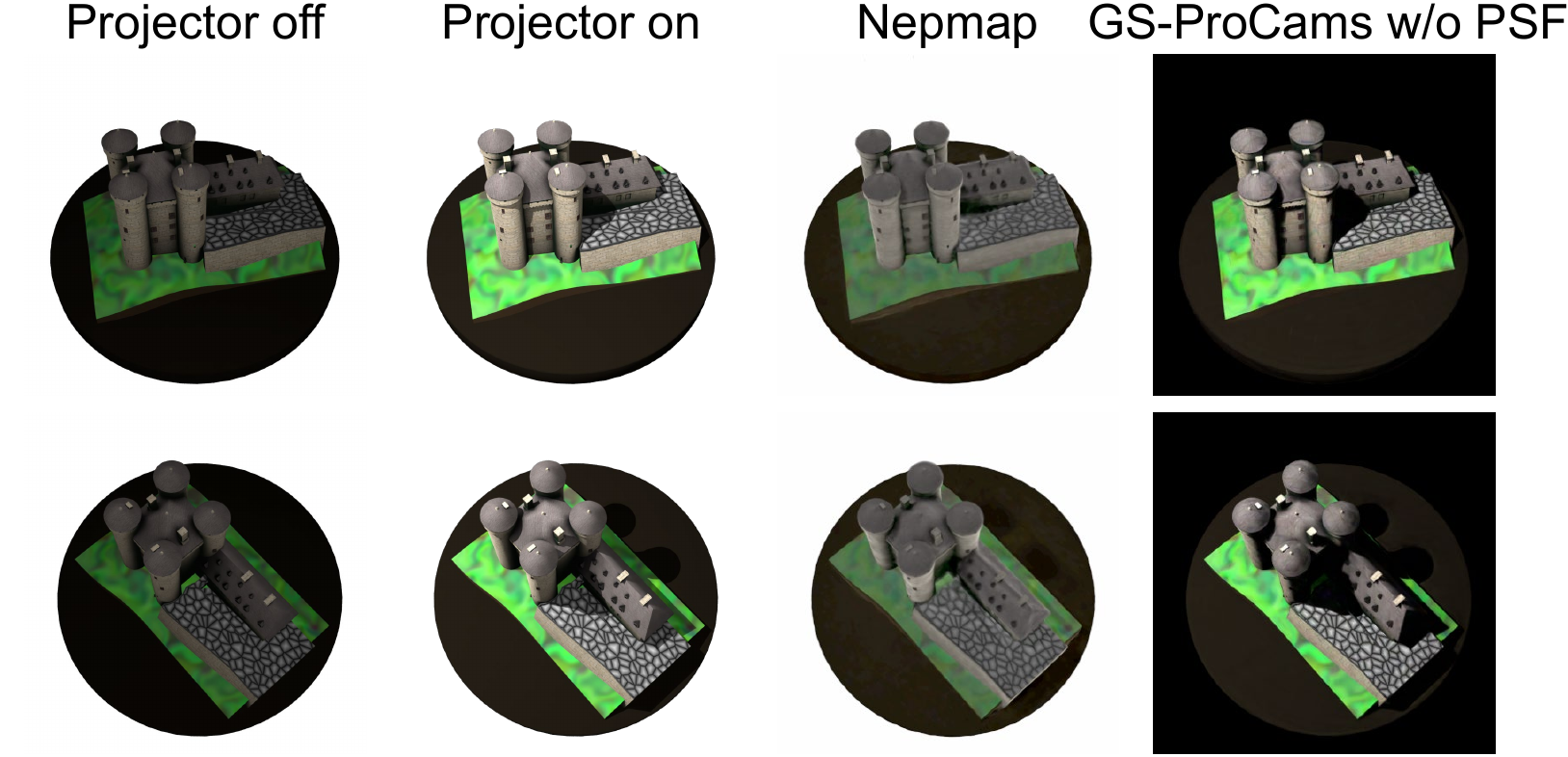}
 \caption{\textbf{Learned albedo on the Nepmap synthetic dataset~\cite{Erel2023Nepmap}.} We present the albedo maps learned by Nepmap and our GS-ProCams under different novel viewpoints of the same scene, shown in two separate rows. The albedo maps are converted from linear to sRGB color space for a better visual experience. The first column shows the scene under camera co-located light only. The second column depicts the same scene under a uniform white projection. GS-ProCams models the co-located lighting as part of the residual color and includes projector occlusions within the albedo estimation.}
 \label{fig: synthetic_scene}
\end{figure}

\subsection{Optimization}
\label{sec:loss}
As shown in \cref{fig: pipeline}, the optimizable parameters associated with each Gaussian point are $\{\mathbf{p}, \mathbf{q}, \mathbf{s}, o, \mathbf{c}_g, \mathbf{b}, r\}$. The optimizable parameters of projector responses are $\{\gamma_p, G_p, \kappa\}$. We impose the following constraints to optimize the GS-ProCams parameters.

\noindent\textbf{Photometric loss.} 
In line with 3DGS~\cite{Kerbl20233DGS}, we consider the combined loss function of $\mathcal{L}_1$ and DSSIM $\mathcal{L}_{\text{DSSIM}}$ to minimize the photometric difference between the simulated result $\hat{\mathbf{I}}_c$ and camera captured ground truth $\mathbf{I}_c$ and set the weight $\lambda=0.2$:
\begin{equation}
\label{eqn:Lrgb}
\mathcal{L}_{c} = (1 - \lambda)\mathcal{L}_1 + \lambda\mathcal{L}_{\text{DSSIM}}
\end{equation}
\noindent\textbf{Geometric regularization.} We also employ depth distortion loss and normal consistency loss~\cite{Huang20242DGS} to encourage the 2D primitives to align with the thin surfaces closely:
\begin{equation}
\label{eqn:Lgeo}
\mathcal{L}_{d} = \sum_{i,j}w_iw_j\left|z_i - z_j\right|, \quad \mathcal{L}_n = \sum_{i}w_i(1 - \mathbf{n}_i^{\top} \mathbf{N}),
\end{equation}
where $\mathbf{n}_i$ is the inherent normal of the intersection between $\mathbf{x}$ and the $i$-th splat.

\noindent\textbf{Materials regularization.} 
In this paper, we jointly estimate surface materials from realistic scenes where roughness maps are anticipated to exhibit smooth characteristics in regions with uniform color~\cite{yao2022neilf}. Consequently, we define a bilateral smoothness term, guided by the albedo, as follows:
\begin{equation}
\label{eqn:Lrm}
\mathcal{L}_\text{material} = {\left\| \nabla{R} \right\|} \text{exp}(-{\left\| \nabla{\mathbf{B}} \right\|}).
\end{equation}
Note that we detach the gradient propagation of albedo $\mathbf{B}$ before computing this loss.

\noindent\textbf{Mask entropy.}
Most projector applications focus only on regions of interest, such as the projector's FOV~\cite{Huang2022CompenNeSt++} or objects~\cite{Erel2023Nepmap}, typically using binary masks to indicate these areas. Therefore, we introduce a cross-entropy constraint~\cite{Wang2021NeuS}, to regularize the optimization for obtaining opaque projection surfaces when $O_m$ is available, and to prune Gaussians outside the region of interest for efficiency. The constraint forces the accumulated opacity $O=\sum_{i \in K}W_i$ to align with $O_m$ and is defined as:
\begin{equation}
\label{eqn:mask}
\mathcal{L}_{\text{entropy}} = -O_m \log O - (1 - O_m) \log (1 - O),
\end{equation}

\noindent\textbf{Total Loss.} Finally, we train GS-ProCams with the following loss function:
\begin{equation}
\label{eqn:Ltotal}
\mathcal{L} = \mathcal{L}_{c} + \lambda_d\mathcal{L}_{d} + \lambda_n\mathcal{L}_n  + \lambda_{m}\mathcal{L}_\text{material} + \lambda_{e}\mathcal{L}_{\text{entropy}},
\end{equation}
where $\lambda_d$, $\lambda_n$, $\lambda_m$, and $\lambda_e$ are the weights for the respective loss terms, selected as 1000, 0.05, 0.002, and 0.1, respectively, based on prior work~\cite{Huang20242DGS, Gao2023Relightable3DGS} for robust optimization. We jointly optimize the GS-ProCams parameters using Adam optimizer~\cite{kingma2014adam} for 20,000 training steps. 

\begin{figure*}[htb]
 \centering 
 \includegraphics[width=\linewidth]{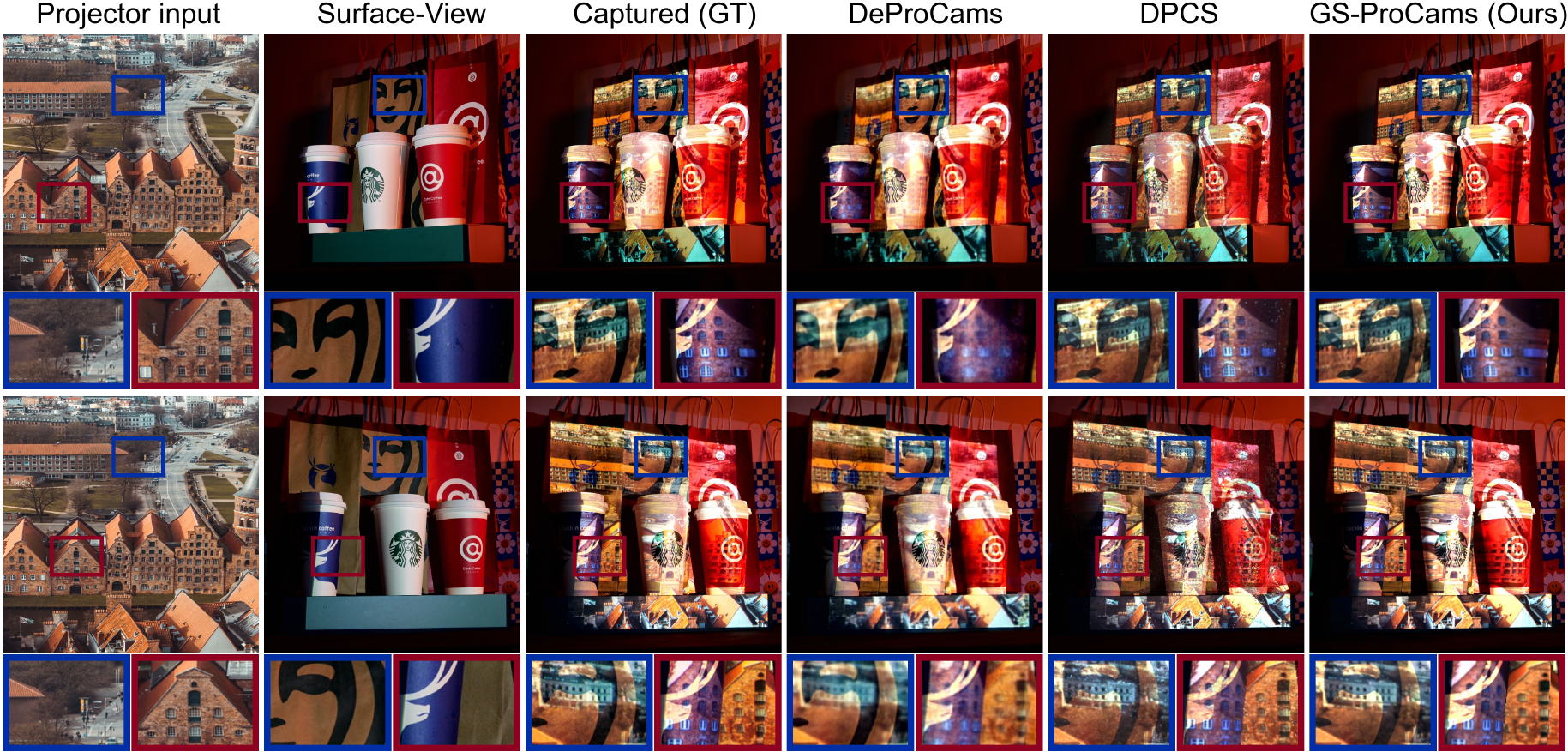}
     \caption{\textbf{Visual comparisons of ProCams simulation on real-world benchmark 
    dataset.} We compare with DeProCams~\cite{Huang2021DeProcams} (view-specific) and DPCS~\cite{Li2025DPCS} (view-specific), for novel projection synthesis. The first row shows a viewpoint included in GS-ProCams training, while the second row shows a novel viewpoint from the same scene that GS-ProCams has never seen during training. The first column shows the same projector input pattern, the second column presents the projection surface under a uniform gray projection and captured in different camera viewpoints, and the remaining columns show the camera-captured ground truth and simulation results. 
    Notably, view-specific methods necessitate repeated data acquisition and training, even for the novel viewpoint, whereas our approach generalizes to novel viewpoints without seeing/retraining on them. 
    Our GS-ProCams consistently produces high-quality simulation results for the novel projection under both trained and novel viewpoints. See supplementary material for more results.}
 \label{fig: relighting}
\end{figure*}

\subsection{Projector compensation}
\label{sec: projector_compensation}
Our GS-ProCams facilitates ProCams simulation through a fully differentiable process, enabling the optimization of projector input patterns to align rendered images with the desired appearances. Once trained, GS-ProCams represents projector compensation as an inverse rendering problem. Given the desired appearance in the camera frame as $\mathbf{I}_{d}$, our framework allows for optimizing a virtual projector input pattern $\hat{\mathbf{I}}_{p}$ to make the simulation result $\hat{\mathbf{I}}_{c}$ in the virtual space align with $\mathbf{I}_{d}$ as follows:
\begin{equation}
\label{eqn: projector_compensation}
\hat{\mathbf{I}}^*_{p} = \arg\min_{\hat{\mathbf{I}}_{p}} \mathcal{L}( \hat{\mathbf{I}}_{c}, \mathbf{I}_{d}),
\end{equation}
where $\hat{\mathbf{I}}^*_{p}$ is the compensated projector input for the desired visual projection appearance $\mathbf{I}_{d}$.

\input{tables/nepmap_synthetic}

\begin{figure}[tb]
 \centering 
 \includegraphics[width=\linewidth]{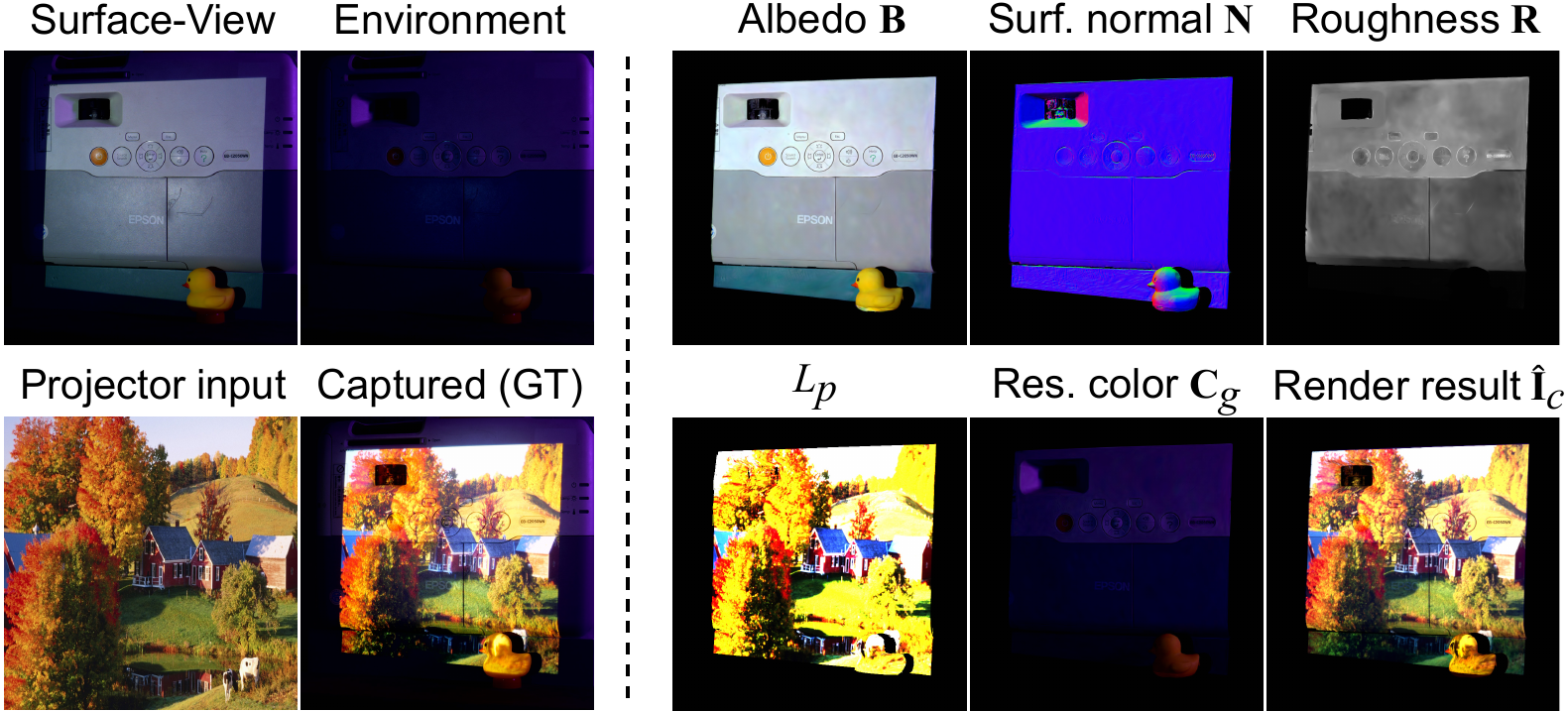}
     \caption{\textbf{Intermediate results.} GS-ProCams explicitly models the ProCams' geometric and photometric mapping using interpretable parameters. On the left, we show the projection surface (Surface-View) and the scene under environment light only (Environment). On the right, are GS-ProCams learned intermediate parameters, ``$L_p$'' is in the linear color space (thus looks overexposed), while ``Albedo'' and ``Residual color'' are converted from linear to sRGB for better visual presentation.}
 \label{fig: relighting_scene}
\end{figure}

\input{tables/procams_simulation}

\input{tables/compensation}

\section{Experiments} 
\label{sec: experiments}
To demonstrate the advantages of our GS-ProCams, we compared it with several state-of-the-art ProCams simulation and/or projector compensation methods~\cite{Erel2023Nepmap, Huang2021DeProcams, Li2025DPCS, Wang2023CompenHR}. In our experiments, we employed Peak Signal-to-Noise Ratio (PSNR), Structural Similarity Index Measure (SSIM), and Learned Perceptual Image Patch Similarity (LPIPS)~\cite{zhang2018lpips} to evaluate the ProCams simulation and projector compensation quality. Additionally, we statistically examined the training time, memory requirements, and inference speed measured in frames per second (FPS) on a workstation equipped with an Intel Core i9-12900K CPU, an Nvidia RTX 3090 GPU, and 64 GB of RAM.
 
\subsection{Synthetic dataset}
\label{sec: Synthetic_dataset}
Our GS-ProCams and the state-of-the-art NeRF-based approach~\cite{Erel2023Nepmap} (\textit{Nepmap} for brevity) are both view-agnostic ProCams simulation methods. Both assume a physically static configuration between the projector and the projection surface, enabling viewpoint-agnostic applications. However, key differences exist. Nepmap embeds the projector into a neural reflectance field~\cite{Boss2021Neural-PIL, Srinivasan2021NeRV, Zhang2021NeRFactor} and relies on a co-located light source under dark-room assumptions to disentangle an object's geometry and material properties. In contrast, GS-ProCams focuses on modeling the projection surface within the projector FOV, requires no co-located lighting, supports environmental illumination, and employs a joint optimization strategy that eliminates the need for multi-stage training, as summarized in \cref{tab:ProCams_works}.

Despite these differences, we compare with Nepmap on its four synthetic scenes using the pre-trained models. Each scene contains 36 novel viewpoints for evaluation, where each viewpoint corresponds to a unique novel projection. Quantitative and qualitative results are shown in \cref{tab: synthetic} and \cref{fig: synthetic}, respectively. Clearly, GS-ProCams outperforms Nepmap in ProCams simulation while showing a significant improvement in efficiency, \ie, about 900 times faster in inference speed and only uses 1/10 of GPU memory for training. Unlike Nepmap, GS-ProCams does not disentangle full-object material properties, but focuses on the projection surface within the projector's FOV. In such scenarios, GS-ProCams interprets co-located lighting as residual color and incorporates projector occlusions into the estimated albedo, as shown in \cref{fig: synthetic_scene}. Moreover, since synthetic data ignores defocus blur, we use a degraded version of GS-ProCams without PSF modeling (\cref{eqn:Lprj}), denoted as \textit{GS-ProCams w/o PSF} for comparison.

\subsection{Real-world benchmark dataset}
\label{sec:Real-world-benchmark-dataset}
We collected a benchmark dataset in real indoor lighting environments. Our projector-camera system consists of a Panasonic Lumix DC-S5 camera and an EPSON CB-965 projector, and an RGB fill light was employed to create various environmental lighting conditions. It is worth noting that, unlike Nepmap~\cite{Erel2023Nepmap}, our method does not require additional light sources to work, the RGB fill light is only used to change ambient lighting conditions. The projector and camera resolutions are both set to $800 \times 800$. Because Nepmap~\cite{Erel2023Nepmap} requires a light source to be co-located with the camera and only works for a dark room as described in \cref{sec: Synthetic_dataset}, it is incompatible with our real-world benchmark setup. Therefore, we exclude it from real-world comparisons. Instead, we compare with three state-of-the-art view-specific methods~\cite{Huang2021DeProcams, Li2025DPCS, Wang2023CompenHR}, using the same projector-camera setup and environmental lighting conditions to ensure a fair comparison. As we will demonstrate, our approach overcomes a fundamental limitation of view-specific baselines, as it generalizes to novel viewpoints while view-specific methods must be retrained for each viewpoint.

We captured 25 training viewpoints per scene for our view-agnostic GS-ProCams. At each viewpoint, we projected one fixed pattern to enhance surface texture for the SfM process~\cite{schoenberger2016sfm}, along with one uniform black and four sampling patterns for model training. These sampling patterns are natural textured images, following~\cite{Huang2021DeProcams}. The plain black projection helps constrain the global illumination residual term by suppressing direct projector influence during joint optimization, while the sampling patterns enable diverse modeling of direct projector effects across views.

Beyond the training set, we collected additional data to evaluate model generalization across novel projections and viewpoints. Specifically, we projected all 100 training patterns onto 5 of the 25 training viewpoints to train view-specific baseline methods. Moreover, we randomly repositioned the camera to capture 8 novel viewpoints beyond the original training set.

At each of these 13 evaluation viewpoints (5 trained + 8 novel), we projected 25 additional novel patterns that were never used during training. Ground-truth geometry was acquired using structured light (SL)~\cite{Moreno2012Calib}, followed by manual cleaning of the point clouds. The resulting geometry is in a relative scale since SfM~\cite{schoenberger2016sfm} performs self-calibration without physical scale references. Projector FOV masks were obtained following~\cite{Huang2021DeProcams}. This dataset establishes a comprehensive benchmark for future research on view-agnostic ProCams simulation under real-world indoor conditions. See supplementary material for more details.

\input{tables/ablation_num_views}

\begin{figure*}[tb]
 \centering 
 \includegraphics[width=\linewidth]{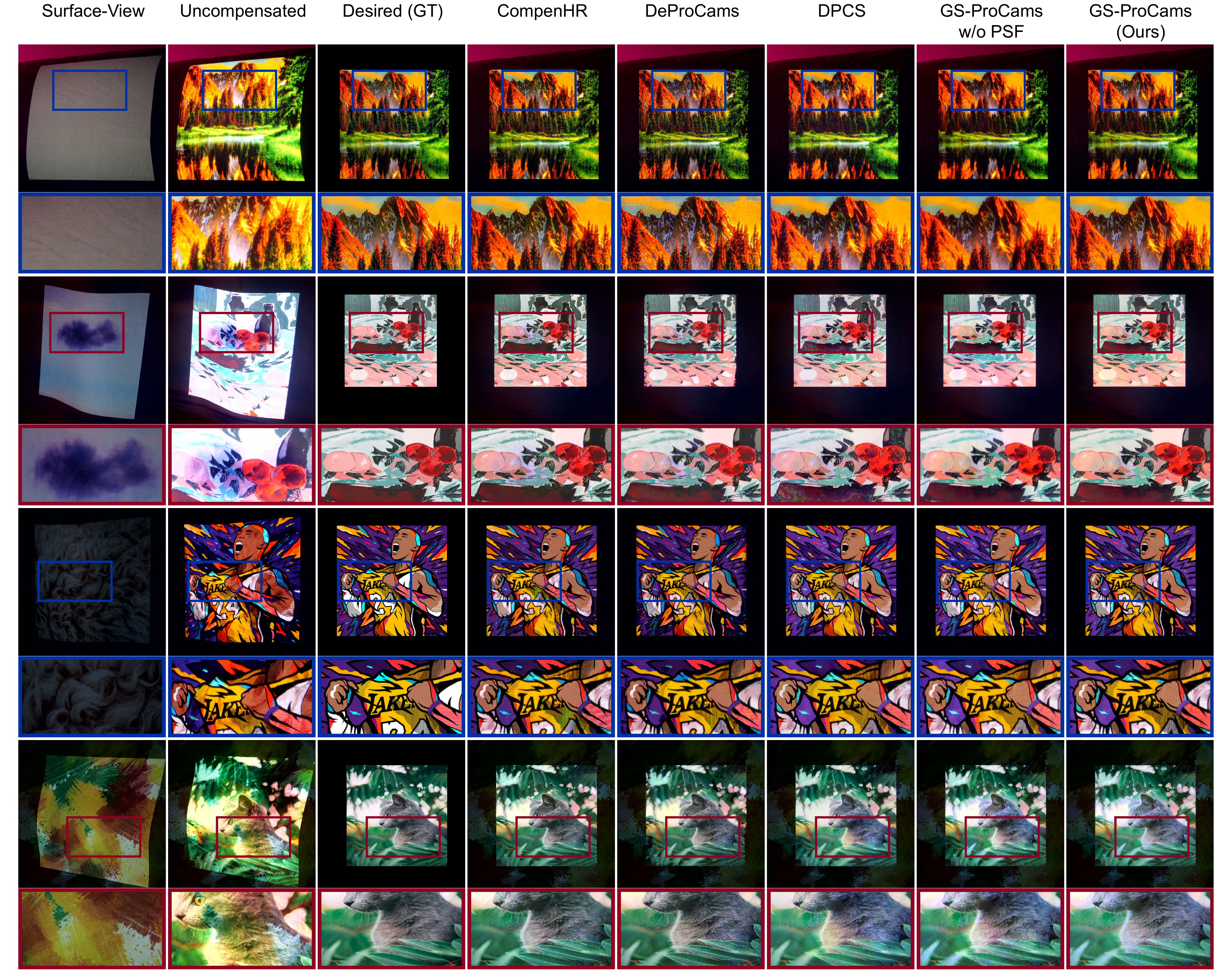}
     \caption{\textbf{Visual comparisons of projector compensation for novel projections.} Each row shows a different surface. The first two columns show the projection surfaces and the uncompensated appearance, respectively. The remaining columns display the target appearance and camera-captured compensated results using different methods. Unlike the three baselines that have seen and been retrained on each viewpoint, our GS-ProCams effectively compensates both geometric and photometric distortions, even if it has not seen or been trained on the four novel viewpoints. See supplementary material for more results.}
 \label{fig: compensation}
\end{figure*}

\begin{figure}[tb]
 \centering 
 \includegraphics[width=\linewidth]{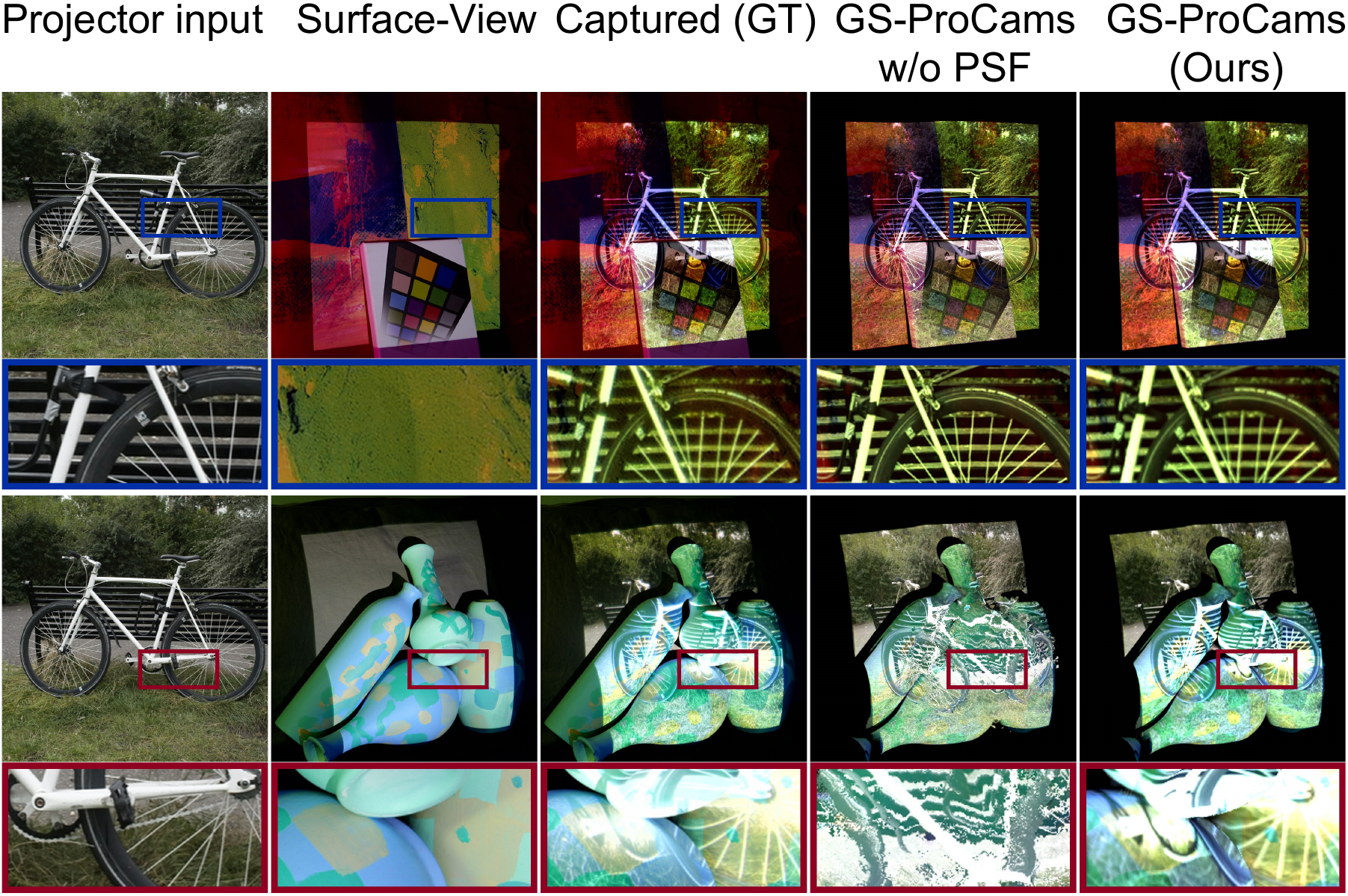}
     \caption{\textbf{Visual comparison w/ and w/o projector PSF estimation.} \textit{GS-ProCams w/o PSF} produces unrealistically sharp projections inconsistent with the ground truth (first row), and fails on complex geometry with large DoF  (second row). In contrast, GS-ProCams yields results that better match real-world observations, enabling more robust ProCams simulation.}
 \label{fig: ablation_psf}
\end{figure}

\vspace{.5em}
\noindent\textbf{ProCams simulation.}
We use two state-of-the-art methods, \textit{DeProCams}~\cite{Huang2021DeProcams} and \textit{DPCS}~\cite{Li2025DPCS} (view-specific), as baseline methods to evaluate our approach (view-agnostic) on the real-world benchmark dataset. We conduct quantitative evaluations by training all models using projector FOV masks for camera viewpoints and computing metrics within the valid SL region, as presented in~\cref{tab: relighting}. Our method outperforms both baselines in synthesizing novel projections, not only at training viewpoints but also under previously unseen novel viewpoints, due to our GS-based formulation. 

We also illustrate qualitative findings with a scene example in \cref{fig: relighting}. We observe that DeProCams may produce blurry results when the projector and viewpoint diverge significantly, potentially due to its reliance on CNNs for shading while simultaneously estimating view-specific per-pixel depth. This poses challenges for high-resolution projections. DPCS, on the other hand, exhibits noise artifacts in some results, likely due to the nature of path tracing~\cite{Li2025DPCS}. Moreover, DPCS relies on a structured light pre-reconstructed surface mesh from a single viewpoint, while GS-ProCams jointly optimizes geometry across multiple viewpoints. While GS-ProCams performs robustly across different viewpoints, it may introduce inconsistencies in regions outside the projection surface, such as shadowed areas and the background, and minor aliasing artifacts along sharp boundaries.
However, GS-ProCams effectively solves the view-agnostic projection mapping problem through physically-based rendering and joint optimization, producing interpretable estimates of the projection surface within the projector FOV from multi-view training images, as shown in~\cref{fig: relighting_scene}.

\vspace{.5em}
\noindent\textbf{Projector compensation.}
We evaluate the practical effectiveness of our framework on projector compensation, with a particular focus on its ability to generalize to unseen novel viewpoints. We compare our method against three state-of-the-art view-specific methods: \textit{CompenHR}~\cite{Wang2023CompenHR}, \textit{DeProCams}~\cite{Huang2021DeProcams}, and \textit{DPCS}~\cite{Li2025DPCS}, all of which require retraining for each specific viewpoint. Since a single projector cannot compensate for regions occluded from the projector direct light, we further evaluate performance on four non-planar, textured surfaces following~\cite{Wang2023CompenHR}. We quantitatively evaluate performance by measuring the difference between the camera-captured compensated results and the desired images, with results in~\cref{tab: compensation} and the visual comparisons in~\cref{fig: compensation}. 

The training time reported in~\cref{tab: compensation} for view-specific methods is the total aggregated time consumed across $V$ novel viewpoints ($V=3$ per scene in our setup), reflecting the repeated, per-viewpoint retraining requirement of these approaches. In contrast, our view-agnostic GS-ProCams reports a single average time per scene, highlighting its immediate readiness for deployment upon initial training. The results demonstrate our method's ability to effectively compensate for both the surface's geometric distortions and photometric interferences at these novel viewpoints without retraining, all while maintaining substantial advantages in overall training time and GPU memory usage, making it a highly efficient and practical solution for real-world applications. 

\subsection{Ablation study}
\noindent\textbf{Simple projector PSF estimation.} We denote the degraded version of GS-ProCams without simple PSF modeling (\cref{eqn:Lprj}) as \textit{GS-ProCams w/o PSF}. In ProCams simulation, we observe that omitting PSF leads to noticeable artifacts on surfaces with complex geometry, as illustrated in~\cref{fig: ablation_psf}, and quantitative results are presented in~\cref{tab: relighting}. For projector compensation, although the absence of PSF does not result in complete failure, it causes noticeable blurring in the results. The quantitative and qualitative results are shown in \cref{tab: compensation} and \cref{fig: compensation}, respectively. This suggests the importance of PSF modeling in ProCams simulation, as discussed in~\cite{Kageyama-Iwai-2022OnlineDeblur,Kusuyama-Iwai-2024Shadow-and-deblur}. While we adopt a simple approximation, better simulation quality may be achieved by considering the insights in these works.

\noindent\textbf{Number of training images.} We evaluated the performance of GS-ProCams under varying numbers of textured training patterns on ProCams simulation. The number of training viewpoints was fixed at 25, and we varied the total number of training images by reducing the number of patterns per viewpoint. Note that GS-ProCams also includes a pure black projection at each training viewpoint, whereas our method uses joint optimization rather than multi-stage training as in Nepmap~\cite{Erel2023Nepmap}. Quantitative results are presented in~\cref{tab: relighting}. The results demonstrate that our method consistently outperforms view-specific baselines across different training pattern numbers. In contrast to the CNN-based method DeProCams~\cite{Huang2021DeProcams}, our method and DPCS~\cite{Li2025DPCS} are both built on explicit modeling of ProCams, and do not require large amounts of training data to achieve high-quality ProCams simulation.

\noindent\textbf{Number of training viewpoints.} We further evaluated the performance of GS-ProCams under a reduced number of training viewpoints (fewer than 25) for ProCams simulation on novel viewpoints. To maintain a total of 100 textured training patterns, we increased the number of training patterns per viewpoint as the number of viewpoints decreased. Quantitative results are presented in \cref{tab: ablation_num_views}. The results demonstrate that GS-ProCams can still achieve high-quality ProCams simulation on unseen viewpoints, even when trained with as few as 4 viewpoints.

\section{Limitations and Future Work}
While our GS-ProCams framework demonstrates state-of-the-art performance in view-agnostic ProCams simulation and projector compensation, it also achieves significantly improved computational efficiency, as demonstrated in our exhaustive experiments (\cref{sec: experiments}). Nonetheless, several limitations remain. In this section, we discuss key challenges and possible directions for future work.

\noindent\textbf{ProCams setup.} Our method is designed for static projector-surface configurations and is not intended for dynamic scenes involving object motion or significant environmental lighting changes.
In this setting, we jointly estimate the geometry and material of the projection surface for efficient view-agnostic projection mapping. As a result, regions outside the projector's field of view, such as shadowed areas, may exhibit undesired artifacts, as accurate estimation in those regions is ill-posed under joint optimization. Moreover, our self-calibration process is based on SfM~\cite{schoenberger2016sfm}, which requires maximizing the camera's coverage of the projection area.

\begin{figure}[tb]
 \centering 
 \includegraphics[width=\columnwidth]{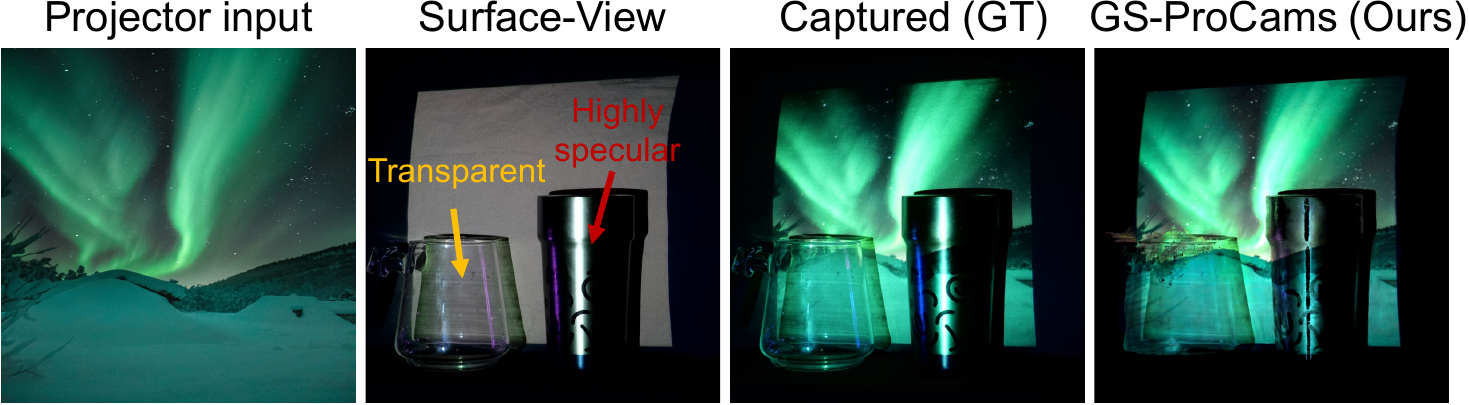}
 \caption{\textbf{Failure case.} GS-ProCams is not capable of handling complex materials such as transparent glass and highly specular surfaces, which exhibit strong view-dependent transmittance and reflections.}
 \label{fig: failure}
\end{figure}

\noindent\textbf{Simplified model.} Real-world scenes are inherently more complex, often involving strong global illumination effects such as inter-reflections and subsurface scattering. We address this in part by jointly estimating a view-dependent residual color term to account for global effects (\cref{eqn: Cres_alpha_blending}), but this remains insufficient. Our method cannot handle highly reflective or transparent materials, such as glass or strongly specular surfaces (\cref{fig: failure}). In addition, the complex characteristics of the hardware system pose further challenges. For example, we estimate a very simple PSF term (\cref{eqn:Lprj}) for defocus blur, as validated in our ablation study (\cref{fig: ablation_psf}). However, for objects with large depth variations, the projector's shallow depth of field (DOF) introduces strong spatially-varying blur that GS-ProCams cannot fully model. Incorporating more accurate PSF estimation approaches~\cite{Kageyama-Iwai-2022OnlineDeblur,Kusuyama-Iwai-2024Shadow-and-deblur} with the GS-ProCams pipeline may better address this issue. 
Another point to note is that we find these factors may lead to ambiguities in both geometry and material parameters estimation, although GS-ProCams still achieves satisfactory ProCams simulation results.

\noindent\textbf{Future work.} To address the above limitations, future work could explore dynamic, view-agnostic ProCams that adapt to object or projector motion as well as lighting changes. Additionally, more precise optical and lighting models may further enhance realism and accuracy.

\section{Conclusion}
We introduce GS-ProCams, the first Gaussian Splatting-based framework for projector-camera systems (ProCams). This framework enables efficient view-agnostic ProCams simulation and projector compensation without retraining. Compared with NeRF-based ProCams, our GS-ProCams simplifies the requirements and supports environmental illumination. While GS-ProCams achieves higher ProCams simulation quality. Moreover, it is 900 times faster and uses only 1/10 of the GPU memory.

\bibliographystyle{abbrv}
\bibliography{ref}

\input{supplementary}

\end{document}

%% file: preamble.tex
\preprinttext{}
\renewcommand{\manuscriptnotetxt}{}
\graphicspath{{figures/}{pictures/}{images/}{./}} 

\usepackage{mathptmx} 
\PassOptionsToPackage{warn}{textcomp} 
\usepackage{textcomp}
\DeclareMathAlphabet{\mathcal}{OMS}{cmsy}{m}{n} 

\usepackage{amsmath}
\usepackage{xspace} 
\usepackage{microtype} 
\usepackage[normalem]{ulem} 
\usepackage{pifont} 

\usepackage{xcolor} 
\usepackage{colortbl}
\usepackage{tabu}
\usepackage{booktabs}

\definecolor{yellow}{RGB}{251, 210, 106}
\definecolor{green}{RGB}{152, 194, 130}
\definecolor{red}{RGB}{230, 117, 101} 
\definecolor{red_darker}{RGB}{204, 64, 64}
\definecolor{green_darker}{RGB}{112, 154, 90}

\newcommand{\bggreen}{\cellcolor{green}}  
\newcommand{\bgred}{\cellcolor{red}} 
\newcommand{\bgyellow}{\cellcolor{yellow}}
\newcommand{\fillcolor}[2]{
  \begingroup
  \setlength{\fboxsep}{1pt}
  \colorbox{#1!80}{\strut #2}
  \endgroup
}

\newcommand{\ie}{i.e.\xspace}

\newcommand{\silentmark}[1]{\raisebox{0pt}[0pt][0pt]{$^{#1}$}}
\newcommand{\best}[1]{\textbf{#1}}
\newcommand{\second}[1]{\underline{#1}}

\makeatletter
\newcommand{\maketitlesupplementary}[1]{
  \twocolumn[{
    \begin{minipage}{\textwidth}
      \begin{center}
        \sffamily\iftoggle{vgtc@journal}{\huge}{\LARGE\bfseries}
        \vgtc@sectionfont
        #1\par
        \vspace{0.5em}
        \large\itshape
        \textemdash\quad Supplementary Materials\quad\textemdash
        \iftoggle{vgtc@journal}{\vspace{14pt}}{\vspace{1\baselineskip}}
      \end{center}
      \vspace{\titlespace}
    \end{minipage}
  }]
  \thispagestyle{empty}
}
\makeatother

%% file: tables/nepmap_synthetic.tex
\begin{table}[tb]
  \caption{\textbf{ProCams simulation results on the Nepmap synthetic dataset~\cite{Erel2023Nepmap}.} Results are averaged over 4 scenes. Since synthetic scenes ignore defocus blur, we use a degraded version of GS-ProCams that excludes the projector PSF for these experiments.}
  \label{tab: synthetic}
  \centering
  \resizebox{\linewidth}{!}{ 
  \begin{tabular}{@{}l @{\hspace{0.1em}} c @{\hspace{0.6em}} c @{\hspace{0.4em}} c @{\hspace{0.4em}} c@{}}
    \toprule
    Method & PSNR$\uparrow$ / SSIM$\uparrow$ / LPIPS$\downarrow$ & Train Time & Train Mem. & Infer FPS \\
    \midrule
    Nepmap~\cite{Erel2023Nepmap} & 27.24 / 0.9445 / 0.061 & $\sim$3 hrs\silentmark{\dag} & $>$24 GB\silentmark{\dag} & 0.3 \\
    GS-ProCams w/o PSF & \best{31.97} / \best{0.9692} / \best{0.036} & 3.6 min & 2.00 GB & 297 \\
    \bottomrule
    \addlinespace[1ex]
    \multicolumn{5}{l}{\silentmark{\dag} Adopted from the original paper}
  \end{tabular}
  } 
\end{table}

%% file: tables/procams_simulation.tex
\begin{table*}[htb]
  \caption{\textbf{ProCams simulation comparisons on a real-world benchmark dataset.} Results are averaged over 7 different scenes. For each scene, we compare our GS-ProCams with baseline methods using 5 training viewpoints and 8 additional novel viewpoints. At each viewpoint, 25 novel projections are evaluated. Note that view-dependent methods must capture additional training data and retrain the model for each single viewpoint, regardless of training or novel viewpoints, while our method only need to be trained on the training viewpoints and can generalize to unseen novel viewpoints without retraining. ``\# Train'' indicates the number of textured patterns used for training each method. $\mathbf{d}_\text{err}$ denotes the mean $L_2$ distance between the inferred and the manually cleaned ground truth point cloud obtained by SL~\cite{Moreno2012Calib}. We use image masks obtained from SL for both inferred and ground truth images to ensure consistency.}
  \label{tab: relighting}
  \centering
  \resizebox{\linewidth}{!}{
  \begin{tabular}{l @{\hspace{2.5em}} l @{\hspace{0.5em}} c @{\hspace{2em}} *{4}{c} c *{4}{c}c}
    \toprule
     &  &  & \multicolumn{4}{c}{\textbf{Trained Viewpoints}} & & \multicolumn{4}{c}{\textbf{Novel Viewpoints}} & \\
    \cmidrule(lr){4-7}
    \cmidrule(lr){9-12}
    \textbf{\# Train} & \textbf{Method} & \textbf{View-agnostic?} & PSNR$\uparrow$ & SSIM$\uparrow$ & LPIPS$\downarrow$& $\mathbf{d}_\text{err}\downarrow$  & & PSNR$\uparrow$ & SSIM$\uparrow$ & LPIPS$\downarrow$ & $\mathbf{d}_\text{err}\downarrow$ \\
    \midrule
    100 & DeProCams~\cite{Huang2021DeProcams} & \textcolor{red_darker}{\ding{55}} & 25.07 & 0.9216 & 0.079 & 0.1106 & & 25.12 & 0.9134 & 0.085 & 0.1264  \\
        & DPCS~\cite{Li2025DPCS} & \textcolor{red_darker}{\ding{55}} & 26.17 & 0.9354 & 0.050 & - &  & 25.87 & 0.9290 & 0.055 & - \\
        & GS-ProCams w/o PSF & \textcolor{green_darker}{\ding{51}} & 24.62 & 0.9258 & 0.062 & 0.0538 &  & 24.14 & 0.9133 & 0.070 & 0.0453 \\
        & GS-ProCams (Ours) & \textcolor{green_darker}{\ding{51}} & \best{26.65} & \best{0.9441}  & \best{0.045} &\best{0.0518} &  & \best{26.10} & \best{0.9335} & \best{0.052} & \best{0.0442} &   \\
    \midrule
    75 & DeProCams~\cite{Huang2021DeProcams} & \textcolor{red_darker}{\ding{55}} & 24.55 & 0.9158 & 0.085 & 0.1248 & & 24.86 & 0.9118 & 0.087 & 0.1401  \\
        & DPCS~\cite{Li2025DPCS} & \textcolor{red_darker}{\ding{55}} & 25.98 & 0.9315 & 0.052 & - &  & 25.73  & 0.9253 & 0.056 & - \\
        & GS-ProCams w/o PSF & \textcolor{green_darker}{\ding{51}} & 24.44 & 0.9239 & 0.063 & 0.0519 &  & 23.96 & 0.9116 & 0.071 & 0.0444 \\
        & GS-ProCams (Ours) & \textcolor{green_darker}{\ding{51}} & \best{26.59} & \best{0.9433} & \best{0.046} & \best{0.0486} &  & \best{26.04} & \best{0.9330} & \best{0.052} & \best{0.0427}  \\
    \midrule
    50 & DeProCams~\cite{Huang2021DeProcams} & \textcolor{red_darker}{\ding{55}} & 24.30 & 0.9143 & 0.086 & 0.1451 & & 24.35 & 0.9074 & 0.090 & 0.1454  \\
        & DPCS~\cite{Li2025DPCS} & \textcolor{red_darker}{\ding{55}} & 26.26 & 0.9355 & 0.050 & - &  & \best{25.98} & 0.9290 & 0.054 & - \\
        & GS-ProCams w/o PSF & \textcolor{green_darker}{\ding{51}} & 24.26 & 0.9227 & 0.064 & 0.0561 &  & 23.75 & 0.9102 & 0.073 & 0.0468 \\
        & GS-ProCams (Ours) & \textcolor{green_darker}{\ding{51}} & \best{26.53} & \best{0.9431} & \best{0.046} & \best{0.0553} &  & \best{25.98} & \best{0.9326} & \best{0.053} & \best{0.0464}  \\
    \midrule
    25 & DeProCams~\cite{Huang2021DeProcams} & \textcolor{red_darker}{\ding{55}} & 23.20 & 0.9063 & 0.094 & 0.1578 & & 23.31 & 0.8997 & 0.099 & 0.1589  \\
        & DPCS~\cite{Li2025DPCS} & \textcolor{red_darker}{\ding{55}} & 26.13 & 0.9339 & 0.052 & - &  & \best{25.84} & 0.9275 & 0.056 & - \\
        & GS-ProCams w/o PSF & \textcolor{green_darker}{\ding{51}} & 23.89 & 0.9177 & 0.067 & \best{0.0549} &  & 23.39 & 0.9051 & 0.076 & \best{0.0474} \\
        & GS-ProCams (Ours) & \textcolor{green_darker}{\ding{51}} & \best{26.25} & \best{0.9400} & \best{0.049} & 0.0559 &  & 25.74 & \best{0.9296} & \best{0.055} & 0.0480  \\
    \bottomrule
  \end{tabular}
  }
\end{table*}

%% file: tables/compensation.tex
\begin{table}[htb]
  \caption{\textbf{Quantitative comparison of projector compensation on novel viewpoints with novel projections.} Results are averaged over 4 scenes. We use \textbf{bold} and \underline{underline} to highlight the best and second-best results respectively. We recorded the training time and the GPU memory required for training. The training time reported for view-specific methods is the total time consumed across $V$ novel viewpoints (where $V$ is the number of novel viewpoints tested, $V=3$ per scene), reflecting the repeated training process required for each viewpoint. In contrast, the time reported for GS-ProCams is the single average time per scene, as it is a view-agnostic method that performs projector compensation on novel viewpoints without seeing or retraining on them.}
  \label{tab: compensation}
  \centering
  \resizebox{\linewidth}{!}{
  \begin{tabular}{@{}l @{\hspace{0.4em}} c @{\hspace{0.6em}} c @{\hspace{0.4em}} c @{\hspace{0.6em}} r @{\hspace{0.4em}} r@{}}
    \toprule
    \textbf{Method} & PSNR$\uparrow$ & SSIM$\uparrow$ & LPIPS$\downarrow$ & Train Time & Train Mem. \\
    \midrule
    CompenHR~\cite{Wang2023CompenHR}\silentmark{\dag} & \best{25.81} & \second{0.8650} & 0.145 & $2.7 \times V$ min & 2.5 GB \\
    DeProCams~\cite{Huang2021DeProcams}\silentmark{\dag} & 23.95 & 0.8364 & 0.167 & $8.8 \times V$ min & 22.8 GB \\
    DPCS~\cite{Li2025DPCS}\silentmark{\dag} & 24.04 & 0.8427 & 0.140 & $8.3 \times V$ min & 6.5 GB \\
    GS-ProCams w/o PSF & 23.70 & 0.8482 & \second{0.135} & 8.8 min & 3.1 GB \\  
    GS-ProCams (Ours) & \second{24.59} & \best{0.8652} & \best{0.121} & 10.5 min & 3.1 GB \\
    \midrule
    Uncompensated & 12.52 & 0.629 & 0.315 & - & - \\ 
    \bottomrule
    \addlinespace[1ex]
    \multicolumn{6}{l}{\small \silentmark{\dag}~View-specific method} 
  \end{tabular}
  }
\end{table}

%% file: tables/ablation_num_views.tex
\begin{table}[tb]
  \caption{\textbf{ProCams simulation results by our GS-ProCams under different numbers of training viewpoints.} We keep the number of textured patterns used for training fixed at 100, while reducing the number of training viewpoints. Note that each viewpoint includes one additional pure black pattern, which also decreases correspondingly. Novel projection synthesis is performed on 8 novel viewpoints. GS-ProCams is capable of maintaining high-quality results even with as few as 4 training viewpoints.}
  \label{tab: ablation_num_views}
  \centering
  \resizebox{\linewidth}{!}{
  \begin{tabular}{@{\hspace{1.5em}} l @{\hspace{2.5em}} c@{\hspace{2em}}c@{\hspace{2em}}c @{\hspace{2.5em}} c @{\hspace{1.5em}}}
    \toprule
    \textbf{\# Train } & PSNR$\uparrow$ & SSIM$\uparrow$ & LPIPS$\downarrow$ & $\mathbf{d}_\text{err} \downarrow$ \\
    \midrule
    25 & 26.10 & 0.9335 & 0.052 & \best{0.0442} \\
    20 & 26.11 & 0.9336 & 0.052 & 0.0457 \\
    10 & \best{26.19} & \best{0.9343} & \best{0.051} & 0.0555 \\
    5  & 25.97 & 0.9319 & 0.052 & 0.0575 \\
    4  & 25.80 & 0.9309 & 0.054 & 0.0566 \\
    2  & 24.25 & 0.9164 & 0.068 & 0.0678 \\
    \bottomrule
  \end{tabular}
  }
\end{table}

%% file: supplementary.tex
\clearpage
\maketitlesupplementary{GS-ProCams: Gaussian Splatting-Based Projector-Camera Systems}
\appendix
\setcounter{page}{1}

\section{Introduction}
In the supplementary material, we provide a detailed description of the real-world data acquisition process in~\cref{sec: data}, further implementation details of GS-ProCams in~\cref{sec: implementation}, and additional results from experiments on ProCams simulation and projector compensation in~\cref{sec:Results}.

\section{Real-world data acquisition}
\label{sec: data}
We use an RGB camera to acquire data in indoor lighting environments. Since a fixed projector illuminates a specific region, we randomly position multiple camera viewpoints along the hemispherical surface surrounding the projection surface and capture data sequentially.

\begin{figure}[tb]
 \centering 
 \includegraphics[width=\columnwidth]{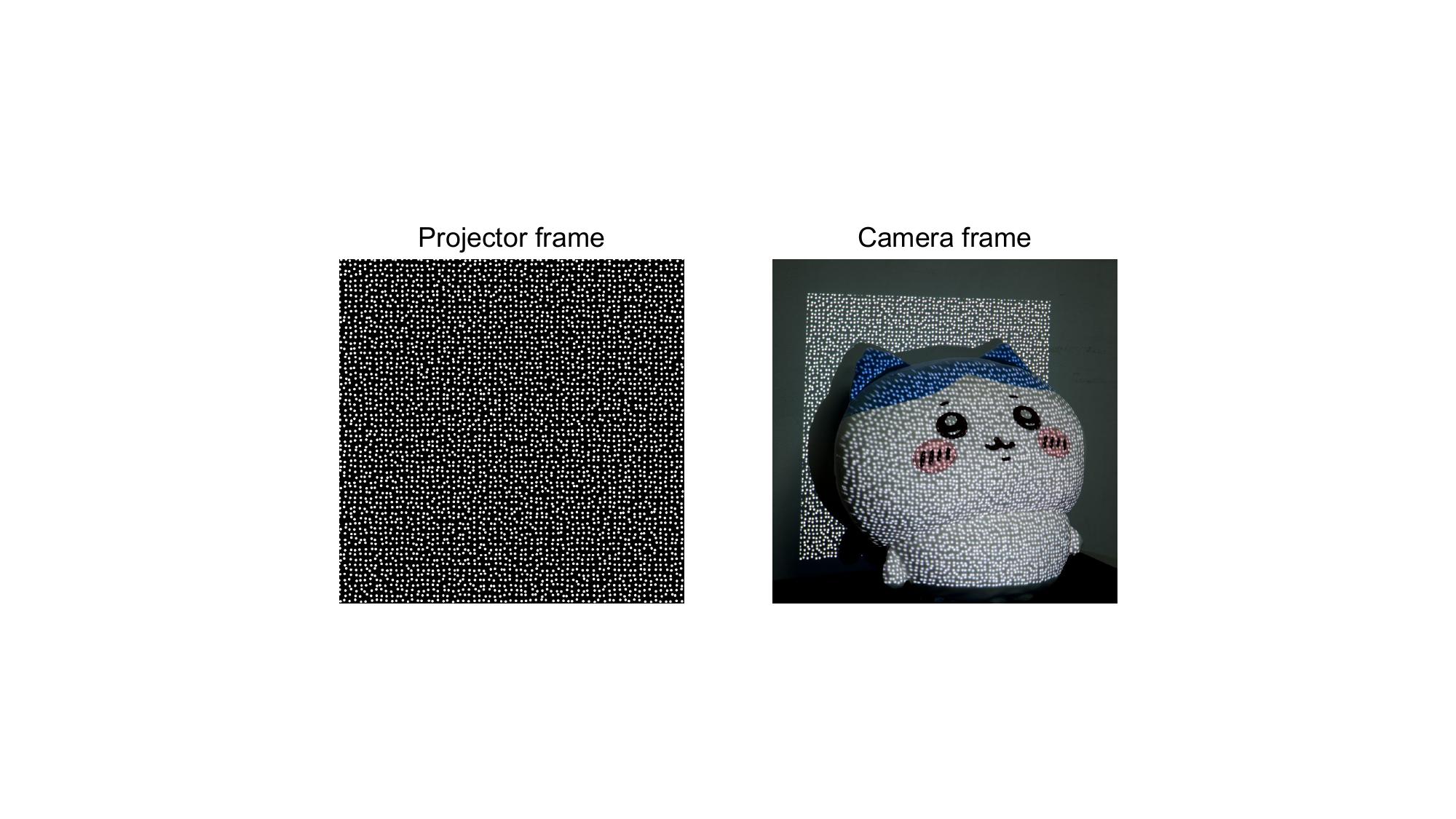}
     \caption{\textbf{Images registration.} We projected a fixed pattern (left) and captured an image using the camera (right) for each viewpoint, thereby determining the positions of the projector and camera viewpoints, using COLMAP~\cite{schoenberger2016sfm}.}
 \label{fig: colmap}
\end{figure}

We employ the pinhole camera model to represent both the projector and the camera viewpoints. A digital speckle pattern as illustrated in \cref{fig: colmap} is projected for each viewpoint, enabling us to obtain the locations and rotations of the projector and the camera viewpoints using COLMAP~\cite{schoenberger2016sfm}. When applying the trained GS-ProCams to a previously unseen viewpoint, we only need to capture an image from this viewpoint with the same projection and then register it into the existing COLMAP model.

\section{Implementation details}
\label{sec: implementation}
\subsection{BRDF modeling}
We use a simplified Disney BRDF model~\cite{burley2012PBRDisney} to approximate the BRDF $f$. Specifically, the BRDF $f$ is composed of a diffuse term $f_{d} = \frac{\mathbf{B}}{\pi}$ and a specular term:
\begin{equation}
\label{eqn:BRDF}
f_{s}(\omega_o, \omega_p) = \frac{DFG}{4 (\mathbf{N} \cdot \omega_p) \cdot (\mathbf{N} \cdot \omega_o)},
\end{equation}
where $D$, $F$, and $G$ denote the microfacet distribution function, Fresnel reflection and geometry factor. Define the half vector between $\omega_o$ and $\omega_p$ as $\mathbf{h}$, the normal distribution function $D$ is evaluated as follows:
\begin{equation}
\label{eqn: D}
D = \frac{R^4}{\pi((\mathbf{N} \cdot \mathbf{h})^2(R^4 -1) + 1)^2},
\end{equation}
and the Fresnel reflection term is approximated as:
\begin{equation}
\label{eqn: F}
F = 0.04 + (1 - 0.04)2^{(-5.55473(\omega_o\cdot\mathbf{h})-6.98316)(\omega_o\cdot\mathbf{h})}.
\end{equation}
Define $k=\frac{(R+1)^2}{8}$, the geometry factor is given by:
\begin{equation}
\label{eqn: g}
G = \frac{(\mathbf{N}\cdot\omega_p)(\mathbf{N}\cdot\omega_o)}{(\mathbf{N}\cdot\omega_p\cdot(1-k)+k)(\mathbf{N}\cdot\omega_o\cdot(1-k)+k)}.
\end{equation}

\subsection{Applications}
As the principal manuscript outlines, our approach constitutes an efficient and fully differentiable framework for ProCams applications. Our compensation framework serves as the foundation for diverse projection mapping applications. 

We present some additional applications of GS-ProCams as shown in the principal manuscript, such as projection-based diminished reality and text-driven projection mapping, which are achieved by integrating GS-ProCams with popular large generative models. Specifically, for projector-based diminished reality, we eliminate objects captured in the camera frame using an inpainting model~\cite{suvorov2022lama} and then apply our compensation method to obtain the projector input pattern, which can act on reality to remove the target objects visually. Similarly, we capture the base appearance of a target from a specific viewpoint, and then we can use text prompts to drive diffusion models~\cite{rombach2022diffusion, hachnochi2023cdc} for text-driven projection mapping. It is precisely because GS-ProCams efficiently models view-agnostic geometric and photometric mappings of ProCams that it can facilitate the application of ProCams in reality.

\section{More results}
\label{sec:Results}
\begin{figure*}[tb]
 \centering
 \includegraphics[width=\textwidth]{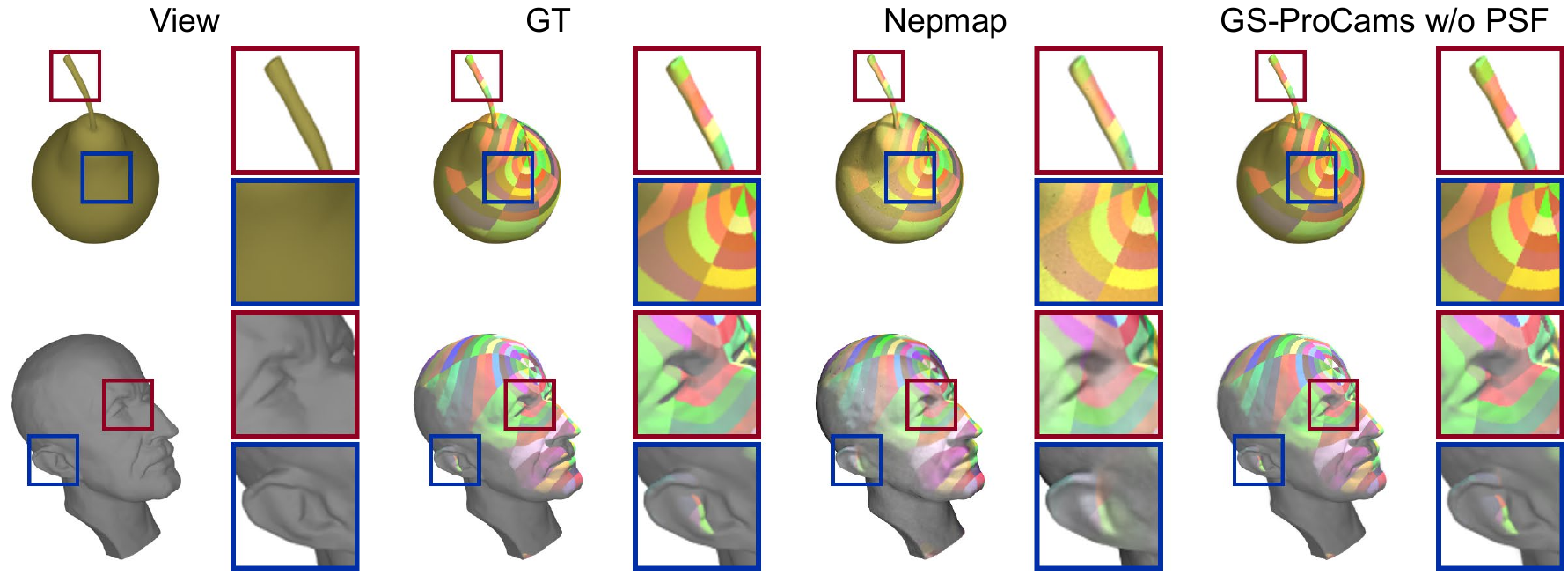}
 \caption{\textbf{Additional visual comparisons of ProCams simulation on the Nepmap synthetic dataset~\cite{Erel2023Nepmap}.} The first column displays an object from a novel viewpoint without projection, the second column shows the object under a novel projection pattern, and the third and fourth columns present the simulated results from two methods.}
 \label{fig: synthetic_sup}
\end{figure*}

\begin{figure*}[tb]
 \centering
 \includegraphics[width=\textwidth]{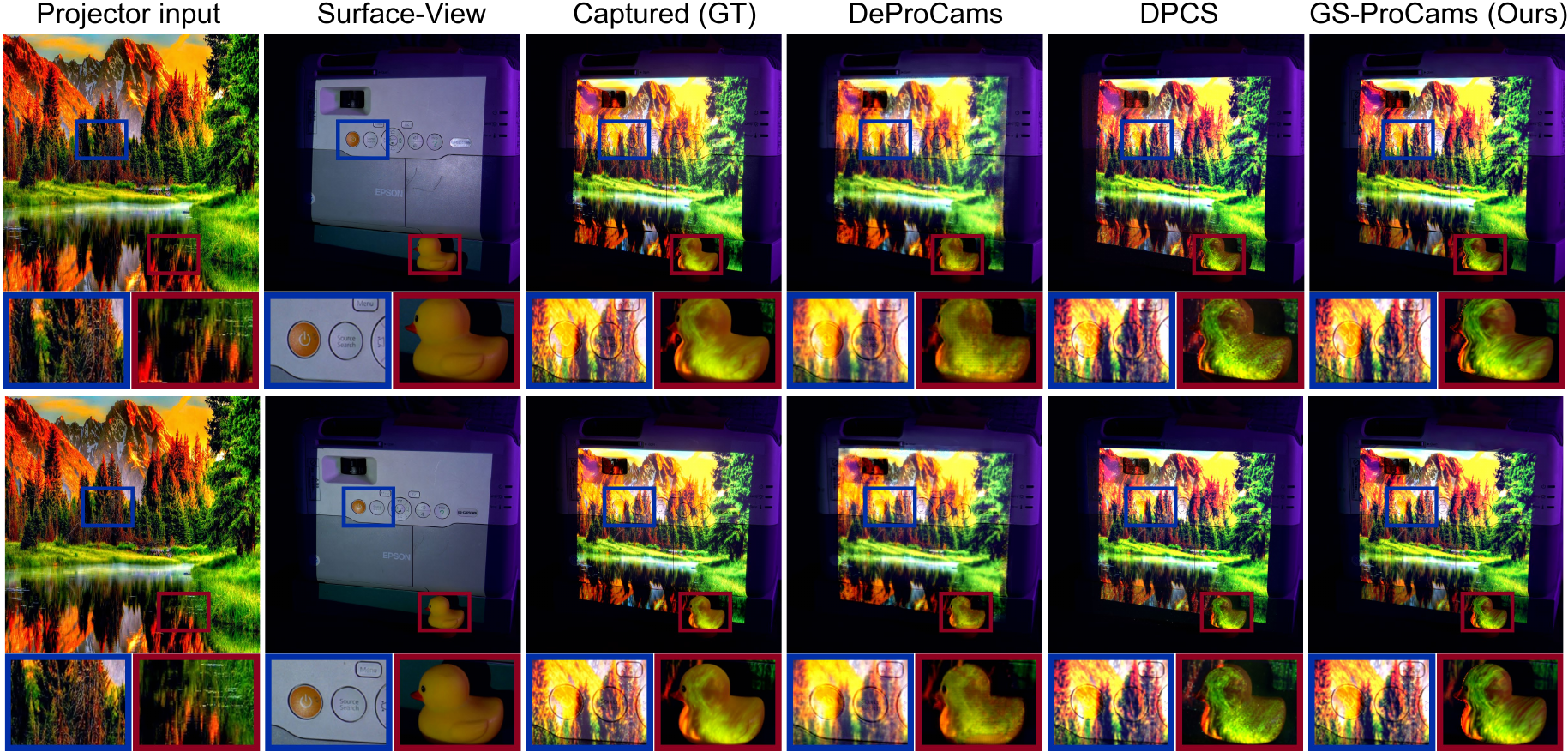}
 \caption{\textbf{Additional visual comparisons of ProCams simulation on real-world benchmark dataset.} The top row shows results from a camera viewpoint used during GS-ProCams training, while the bottom row shows results from a novel viewpoint that GS-ProCams has not seen during training. The first column shows the projector input pattern, followed by the projection surface under uniform gray projection. The remaining columns present the ground truth and simulation results from each method.} 
 \label{fig: relighting_sup}
\end{figure*}

\noindent\textbf{Procams simulation.}
We present the quantitative per-scene results on the Nepmap synthetic dataset~\cite{Erel2023Nepmap} in~\cref{tab: synthetic_sup}. Additional visual comparisons of ProCams simulation on the synthetic dataset are shown in~\cref{fig: synthetic_sup}. Additional visual comparisons of ProCams simulation on real-world benchmark dataset are given in~\cref{fig: relighting_sup}. 

\begin{figure*}[tb]
 \centering
 \includegraphics[width=\textwidth]{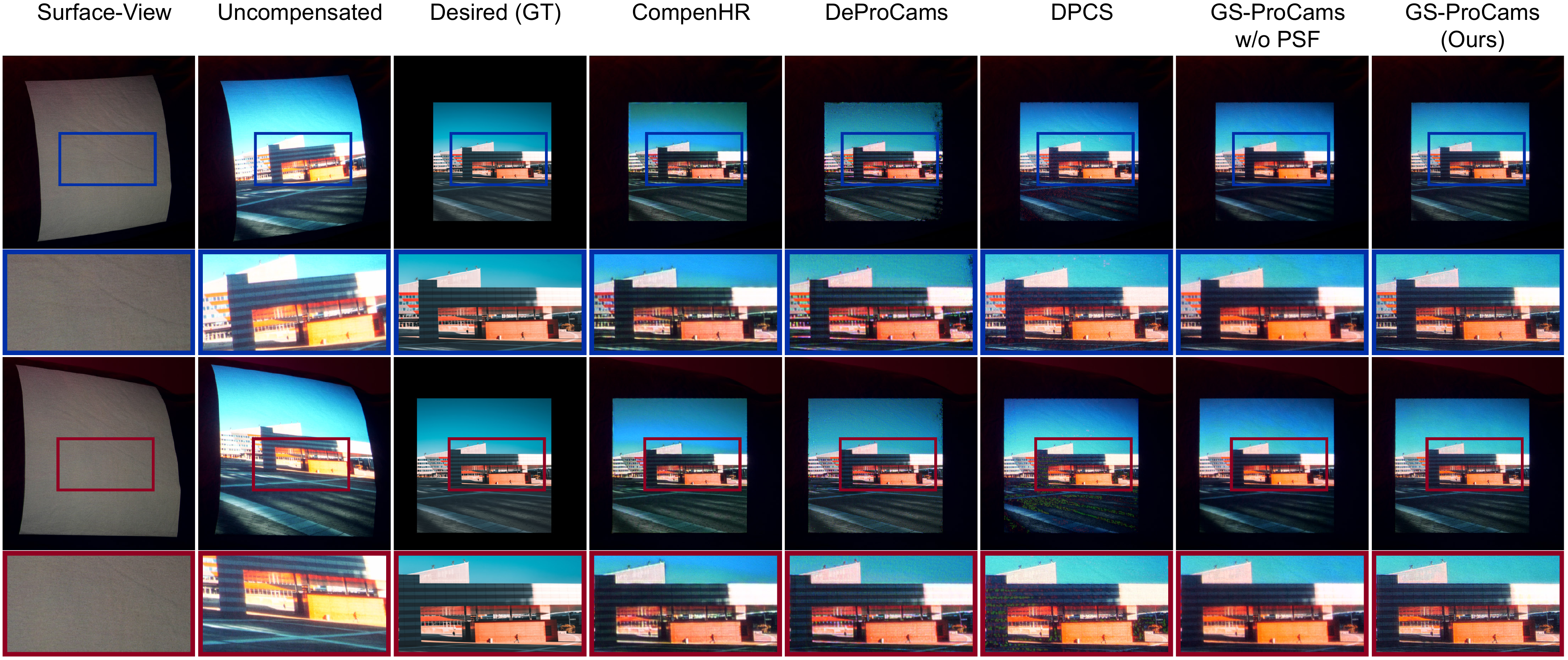}
 \caption{\textbf{Visual comparisons of projector compensation across novel viewpoints.} Each row presents a different viewpoint of the same surface. The first three columns show the projection surface, the uncompensated appearance, and the desired appearance. The remaining columns display the camera-captured compensated results for the same projection from different methods. While view-specific methods require additional data acquisition and retraining for each novel viewpoint, GS-ProCams generalizes effectively, delivering consistent and high-quality results across unseen views.}
 \label{fig: sup_compensation_pillow}

 \vspace{1.5em}

 \includegraphics[width=\textwidth]{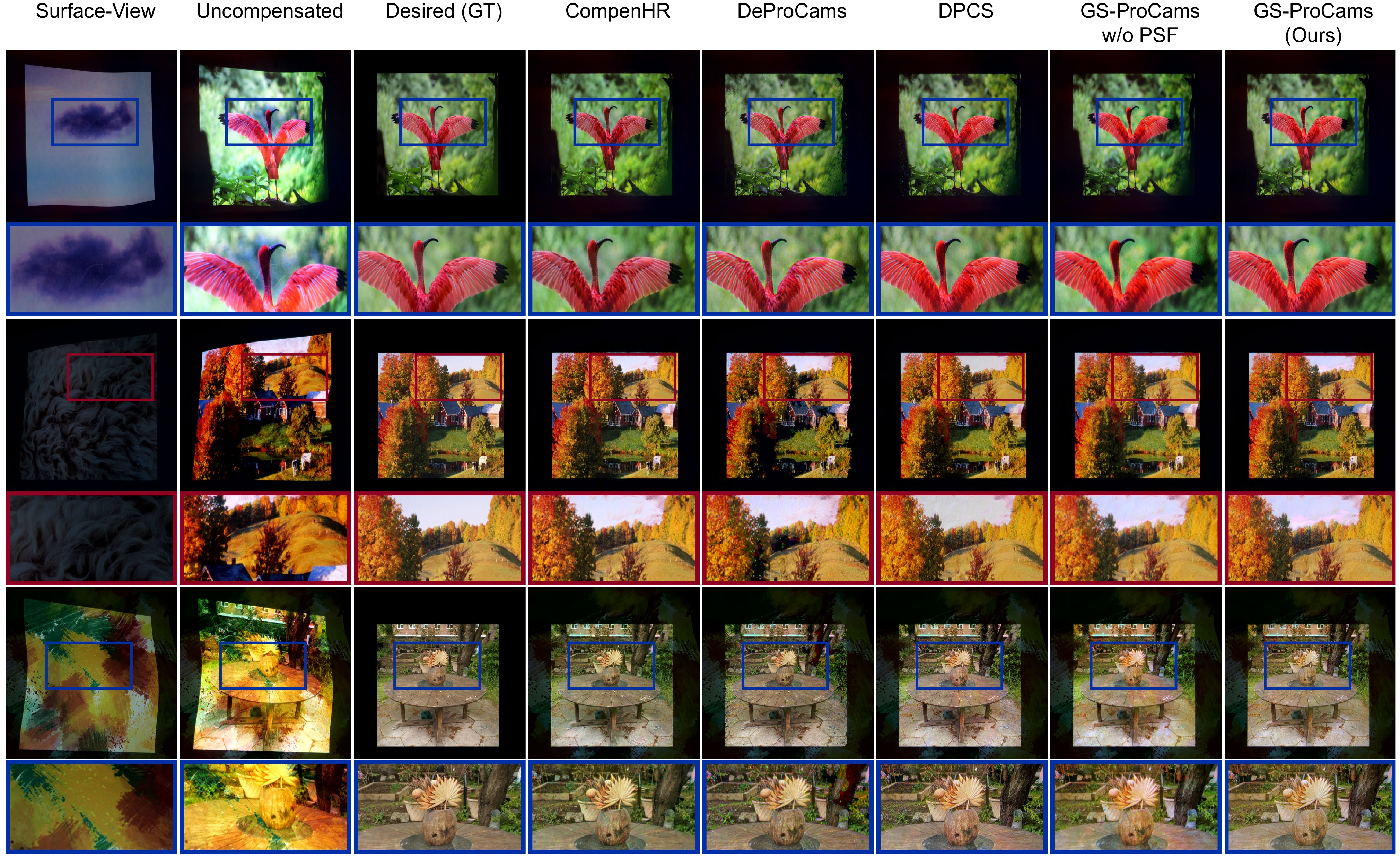}
 \caption{\textbf{Additional visual comparisons of projector compensation.} Each row illustrates the results of a scene under a novel viewpoint that our proposed GS-ProCams has not previously seen. The other three view-specific methods necessitate repeated data capturing and training for these viewpoints. The first three columns respectively illustrate the projection surface from the specified viewpoint, the uncompensated appearance, and the expected appearance. The remaining columns display the compensated results from different models.}
 \label{fig: sup_compensation_3s}
\end{figure*}

\vspace{.5em}
\noindent\textbf{Projector compensation.}
To further evaluate the reliability of GS-ProCams compensation under novel viewpoints, we test it on the same scene using the same projector input across different camera views, as illustrated in~\cref{fig: sup_compensation_pillow}. Additional comparisons of projector compensation on different surfaces are provided in~\cref{fig: sup_compensation_3s}.

\begin{table*}[tb]
  \caption{\textbf{ProCams simulation results on the Nepmap synthetic dataset~\cite{Erel2023Nepmap}.} Since synthetic scenes ignore defocus blur, we use a degraded version of GS-ProCams that excludes the projector PSF for these experiments.}
  \label{tab: synthetic_sup}
  \centering
  \resizebox{\linewidth}{!}{
  \begin{tabular}{@{}l @{\hspace{0.1em}} c @{\hspace{0.6em}} c @{\hspace{0.6em}} c @{\hspace{0.6em}} c @{\hspace{0.6em}} c @{\hspace{0.4em}} c @{\hspace{0.4em}} c@{}}
    \toprule
     & \textbf{Zoo} & \textbf{Castle} & \textbf{Pear} & \textbf{Planck} & \multicolumn{3}{c}{\textbf{Average}} \\
    \cmidrule(lr){2-2}
    \cmidrule(lr){3-3}
    \cmidrule(lr){4-4}
    \cmidrule(lr){5-5}
    \cmidrule(lr){6-8}
    \textbf{Method} & PSNR$\uparrow$ / SSIM$\uparrow$ / LPIPS$\downarrow$ & PSNR$\uparrow$ / SSIM$\uparrow$ / LPIPS$\downarrow$ & PSNR$\uparrow$ / SSIM$\uparrow$ / LPIPS$\downarrow$ & PSNR$\uparrow$ / SSIM$\uparrow$ / LPIPS$\downarrow$ & Train Time & Train Mem. & Infer FPS \\
    \midrule
    Nepmap~\cite{Erel2023Nepmap} & 25.68 / 0.9300 / 0.046 & 28.93 / 0.9404 / 0.065 & 27.29 / 0.9648 / 0.062 & 27.06/ 0.9427 / 0.071 & $\sim$3 hrs\silentmark{\dag} & $>$24 GB\silentmark{\dag} & 0.3 \\
    GS-ProCams w/o PSF & \best{27.77} / \best{0.9527} / \best{0.044} & \best{34.96} / \best{0.9751} / \best{0.031} & \best{33.90} / \best{0.9800} / \best{0.031} & \best{31.25} / \best{0.9689} / \best{0.040} & 3.6 min & 2.00 GB & 297 \\
    \bottomrule
    \addlinespace[1ex]
    \multicolumn{5}{l}{\silentmark{\dag} Adopted from the original paper}
  \end{tabular}
  }
\end{table*}

